\documentclass[review]{elsarticle}

\usepackage{lineno}
\usepackage[textsize=small]{todonotes}
\usepackage{color}
\usepackage{microtype}

\usepackage{graphicx}
\usepackage{amsmath}
\usepackage{amsfonts}
\usepackage{mathtools}

\DeclarePairedDelimiter{\floor}{\lfloor}{\rfloor}

\usepackage{tabulary}
\usepackage{makecell}

\newcommand{\giannis}[1]{\textcolor{blue}{#1}}

\usepackage[textsize=small]{todonotes}

\newcommand{\etal}{et al.\ }  

\newcommand{\eg}{e.g., }
\newcommand{\ie}{i.e., }

\newcommand{\Figref}[1]{Figure~\ref{#1}}  
\newcommand{\tabref}[1]{Table~\ref{#1}}

\newcommand{\secref}[1]{Section~\ref{#1}}
\newcommand{\equref}[1]{(\ref{#1})}

\usepackage[colorlinks]{hyperref}

\hypersetup{colorlinks = true,linkcolor = blue,anchorcolor =red,citecolor = blue,filecolor = red,urlcolor = red,pdfauthor=author}

\biboptions{authoryear}

\makeatletter
\def\ps@pprintTitle{%
  \let\@oddhead\@empty
  \let\@evenhead\@empty
  \def\@oddfoot{\reset@font\hfil\thepage\hfil}
  \let\@evenfoot\@oddfoot
}
\makeatother

\usepackage{tikz,xcolor,hyperref}

\definecolor{lime}{HTML}{A6CE39}
\DeclareRobustCommand{\orcidicon}{
	\begin{tikzpicture}
	\draw[lime, fill=lime] (0,0) 
	circle [radius=0.16] 
	node[white] {{\fontfamily{qag}\selectfont \tiny ID}};
	\draw[white, fill=white] (-0.0625,0.095) 
	circle [radius=0.007];
	\end{tikzpicture}
	\hspace{-2mm}
}

\foreach \x in {A, ..., Z}{\expandafter\xdef\csname orcid\x\endcsname{\noexpand\href{https://orcid.org/\csname orcidauthor\x\endcsname}
			{\noexpand\orcidicon}}
}

\usepackage{etoolbox}
\patchcmd{\pprintMaketitle}
 {\ifvoid\absbox\else\unvbox\absbox\par\vskip10pt\fi}
 {\ifvoid\absbox\else\clearpage\unvbox\absbox\par\vskip30pt\fi}
 {}{}
\patchcmd{\pprintMaketitle}
 {\hrule\vskip12pt}
 {}
 {}{}
\patchcmd{\pprintMaketitle}
 {\hrule\vskip12pt}
 {}
 {}{}
\appto{\pprintMaketitle}{\clearpage}

\begin{document}

\begin{frontmatter}

\title{Graph Convolutional Neural Networks with Node Transition Probability-based Message Passing and DropNode Regularization}

\author[vub,imec]{Tien Huu Do\corref{cor1}}
\ead{thdo@etrovub.be}
\author[vub,imec]{Duc Minh Nguyen}
\ead{mdnguyen@etrovub.be}
\author[vub,imec]{Giannis Bekoulis}
\ead{gbekouli@etrovub.be}
\author[vub]{Adrian Munteanu}
\ead{acmuntea@etrovub.be}
\author[vub,imec]{Nikos Deligiannis\orcidA{}}
\ead{ndeligia@etrovub.be}
\cortext[cor1]{Corresponding author}
\address[vub]{Vrije Universiteit Brussel, Pleinlaan 2, B-1050 Brussels, Belgium }
\address[imec]{imec, Kapeldreef 75, B-3001 Leuven, Belgium}

\begin{abstract}
  Graph convolutional neural networks (GCNNs) have received much attention recently, owing to their capability in handling graph-structured data. 
Among the existing GCNNs, many methods can be viewed as instances of a neural message passing motif; features of nodes are passed around their neighbors, aggregated and transformed to produce better nodes' representations. 
Nevertheless, these methods seldom use node transition probabilities, a measure that has been found useful in exploring graphs. 
Furthermore, when the transition probabilities are used, their transition direction is often improperly considered in the feature aggregation step, resulting in an inefficient weighting scheme. 
In addition, although a great number of GCNN models with increasing level of complexity have been introduced, the GCNNs often suffer from over-fitting when being trained on small graphs. Another issue of the GCNNs is over-smoothing, which tends to make nodes' representations indistinguishable. 
This work presents a new method to improve the message passing process based on node transition probabilities by properly considering the transition direction, leading to a better weighting scheme in nodes' features aggregation compared to the existing counterpart. 
Moreover, we propose a novel regularization method termed \textit{DropNode} to address the over-fitting and over-smoothing issues simultaneously. 
DropNode randomly discards part of a graph, thus it  creates multiple deformed versions of the graph, leading to data augmentation regularization effect. 
Additionally, DropNode lessens the connectivity of the graph, mitigating the effect of over-smoothing in deep GCNNs. 
Extensive experiments on eight benchmark datasets for node and graph classification tasks demonstrate the effectiveness of the proposed methods in comparison with the state of the art. 
\end{abstract}

\end{frontmatter}


\section{Introduction}
\label{sec:introduction}
In the recent years, we have witnessed an increasing interest in deep learning-based techniques for solving problems with data living in non-Euclidean domains, i.e., graphs and manifolds. In order to handle data in such domains, graph neural networks (GNNs) have been proposed. 
There are several types of GNNs, however recently graph convolutional neural networks (GCNNs) have attracted a great deal of attention owing to their state-of-the-art performance in a number of tasks including text~\citep{yao2019graph} and image~\citep{quek2011structural} classification, semantic segmentation~\citep{qi20173d}, question answering~\citep{teney2017graph} and matrix completion~\citep{do2019matrix}. 
There are two main approaches for GCNNs, namely, \textit{spectral} and \textit{spatial}~\citep{zhou2018graph} approaches. The former refers to models which define convolution operators based on graph Fourier transform~\citep{defferrard2016convolutional,henaff2015deep,bruna2013spectral,ortega2018graph}. On the other hand, models that follow the latter approach directly generalize the classical convolution operator to non-Euclidean domains (\ie by spatially aggregating information over nodes' neighborhoods in the graphs~\citep{hamilton2017inductive,velivckovic2017graph,niepert2016learning,gao2018large,atwood2016diffusion}). The spatial approach has become more popular lately due to its simplicity, lower computational cost and better scalability to large graphs.

In general, the operations of most GCNN models can be expressed in two phases, namely, the (\textit{i}) \textit{Message Passing} and (\textit{ii}) \textit{Readout} phases. 
In phase (\textit{i}), a node in the graph receives messages from its neighbors, where each message is computed either from features of a neighboring node or features of an edge attached to the node. Two steps that are often performed in the message passing phase are the \textit{Aggregate} and \textit{Update} steps~\citep{zhou2018graph}. In the Aggregate step, a node gathers messages from its neighbors by performing the weighted sum operation~\citep{kipf2016semi,hamilton2017inductive,zhang2018} or employing a long short-term memory neural network (LSTM)~\citep{hamilton2017inductive}. 
In the Update step, a non-linear transformation is then applied on the aggregated messages to produce the updated node representations. 
Common transformations used in the Update step include pre-defined functions, such as sigmoid and ReLU~\citep{bruna2013spectral,defferrard2016convolutional}, or learnable functions, such as gated recurrent units (GRU)~\citep{li2015gated} and fully connected neural networks~\citep{kearnes2016molecular,battaglia2016interaction,hamilton2017inductive,duvenaud2015,kipf2016semi,zhang2018}. 
In GCNN models, the Message Passing phase is often implemented using \textit{graph convolutional layers}. By stacking multiple graph convolutional layers, one can build a deep GCNN model and enable (\textit{a}) passing messages between nodes that are more than one hop apart (\ie nodes connected via other nodes)
and (\textit{b}) learning abstract intermediate node embeddings through a series of non-linear transformations. 
In phase (\textit{ii}), the learned node embeddings are converted to the final representations for the nodes or for the whole graph. These representation are then fed to a classifier to produce suitable outputs depending on downstream tasks.

In GCNN models, the non-Euclidean characteristics of the data are captured mainly in the Message Passing phase. As such, designing an effective message passing scheme is of high importance, however it remains a challenging task. 
To address this challenge, various message passing formulations have been proposed~\citep{gilmer2017neural,duvenaud2015,kipf2016semi,hamilton2017inductive}. 
Recently, formulations based on node transition probabilities have been proven beneficial for multiple graph-based classification tasks~\citep{atwood2016diffusion,zhang2018}. Nevertheless, such formulations have not received enough attention from the research community. 
In addition, other challenges such as the \textit{over-fitting} and \textit{over-smoothing} problems (\eg especially when deploying \textit{deep} GCNNs) often arise. 
To mitigate over-fitting, popular regularization methods such as $\ell_1$, $\ell_2$ and \textit{dropout}~\citep{srivastava2014dropout} are usually employed. 
Nevertheless, performance gains brought by these methods generally diminish as GCNNs become deeper~\citep{kipf2016semi}. 
It is worth noting that over-fitting is general, i.e., it may happens for different types of neural networks, including GCNNs. 
On the other hand, over-smoothing~\citep{li2018deeper,chen2019measuring} is specific for deep GCNNs as a result of the inherent smoothing effect of these models. 
Specifically, when many graph convolutional layers are stacked together, the obtained representations of nodes in different classes (or clusters) become indistinguishable and inseparable, leading to degraded performance in downstream tasks such as node classification.

In this work, we aim to develop effective GCNN models to handle graph-structured data by addressing the aforementioned challenges. 
Firstly, we design a GCNN message passing scheme employing the transition-based approach.  
Specifically, we observe that the transition directions employed in existing works~\citep{atwood2016diffusion,zhang2018} are sub-optimal since high weights (\ie influence) are assigned to ``popular'' nodes (\ie the ones with multiple connections). Based on this observation, 
we propose a novel message passing scheme that eliminates the aforementioned issue by assigning balanced weights to all the nodes in the graph. We experimentally validate the effectiveness of the proposed scheme in multiple graph classification tasks. In addition, we introduce a novel method called \textit{DropNode} to address the over-fitting and over-smoothing problems on GCNNs. The idea behind this method is that the underlying structure of the graph is modified during training by randomly sub-sampling nodes from the graph. Despite its simplicity, the DropNode method is highly effective, especially in training deep GCNN models. 

To summarize, our contributions in this paper are three-fold:
\begin{itemize}
    \item We propose a novel neural message passing scheme for GCNNs, which leverages the node transition probabilities and also takes the ``popularity'' of  nodes into account;
    \item We propose DropNode, a simple and effective regularization method which is applicable to different GCNN models;
    \item We carry out comprehensive experiments on eight real-world benchmark datasets on both node and graph classification tasks to evaluate the proposed models. Experimental results showcase that our models are able to obtain improved performance over state-of-the-art models.
\end{itemize} 

The remainder of the paper is organized as follows. Section~\ref{sec:related_work} reviews the related work on GCNNs and the commonly-used regularization methods. Section~\ref{sec:preliminaries} introduces the notation and problem formulation, while Section~\ref{sec:method} describes in detail the proposed method. We present the experimental results of the proposed models in Section~\ref{sec:experiment} and we conclude our work in Section~\ref{sec:conclusion}.

\section{Related Work}
\label{sec:related_work}
\subsection{Neural Message Passing}
Most existing GCNN models perform one or multiple \textit{message passing} steps, in which nodes' or edges' features are aggregated following specific message passing schemes~\citep{gilmer2017neural}. Early works, which follow the spectral approach, aggregate messages by summing over transformed features of the nodes where the transformations are parameterized by the eigenvectors of the Laplacian graph~\citep{bruna2013spectral,defferrard2016convolutional}. As these eigenvectors  correspond to the Fourier bases of the graph, each message passing step can be seen as a convolution operation on the features of the graph. More recently, Kipf \etal\citep{kipf2016semi} proposed the Graph Convolutional Network (GCN) model which aggregates node features by employing the normalized graph adjacency matrix as a transformation matrix (in this case, the transformation is a linear projection). Since the message passing scheme of the GCN (in its final form) does not rely on the Fourier analysis of the graph, GCN can be seen as a spatial-based model. Following GCN~\citep{kipf2016semi}, various models based on the spatial approach have been recently proposed. For instance, messages are aggregated by averaging nodes' features in~\citep{hamilton2017inductive}, summing over concatenated features of nodes and edges in~\citep{duvenaud2015}, or by applying neural networks on nodes' features~\citep{hamilton2017inductive} and edges' features in~\citep{battaglia2016interaction,schutt2017quantum}. Most of these spatial-based models utilize the connections between nodes in the graph directly as weights when aggregating node and edge features. The weight of the connections can be pre-computed, or can be learned by leveraging the attention mechanisms, as in the attention graph neural networks (GAT) model~\citep{velivckovic2017graph}. In addition, other models such as the node transition-based ones employ transition probabilities between nodes, where these probabilities are calculated based on random walk procedures~\citep{atwood2016diffusion,zhang2018}. 

Our message passing scheme is inspired by the node transition-based models [see~\citep{atwood2016diffusion,zhang2018}]. The key difference between the proposed method and these works is that we do not overestimate the influence of the popular nodes, which leads to the performance improvement on several graph classification tasks.

\subsection{Graph Regularization Methods}
As the second contribution of this work is DropNode, a regularization technique for GCNNs, this section is dedicated to review existing regularization methods for GCNNs. Additionally, the relations between our method and the existing methods are presented. 

Dropout~\citep{srivastava2014dropout} is a commonly used regularization technique for various types of neural networks, including GCNNs. 
In dropout, a percentage of neural units is kept active during training using Bernoulli distribution, which in turn reduces the co-adapting effect of neural units. In testing, by using the full architecture, the network is able to approximate the averaging of many thinned models, leading to better performance. 
Unlike dropout, our method (DropNode) randomly removes nodes of the underlying graphs during the training of GCNNs. Conceptually, DropNode alters the data instead of neural units of network architecture. 
DropNode is similar to the work of Chen \textit{et al.}~\citep{chen2018fastgcn} in terms of random node sampling. 
However, we take a further step by adding an  \textit{upsampling} operation, which recovers the dropped nodes from the previous sampling step. 
This arrangement allows us to yield the data augmentation effect following the U-Net principle~\citep{ronneberger2015u}. 
In the model proposed by Gao \textit{et al.}~\citep{gao2019graph}, pooling and unpooling operations are used to create a graph-based U-Net architecture. However, the pooling uses \texttt{top-k} selection based on scores of nodes; the scores are computed using a scalar projection of node features. 
This is different from our method as we randomly sample nodes using the Bernoulli distribution. 
As a result, the model in~\citep{gao2019graph} does not have the regularization effect because the architecture is fixed across training and testing. 
In the work of~\citep{rong2019dropedge}, edges are randomly removed to regularize GCNNs, which is similar to our method in terms of modifying the underlying graph. However, as the number of nodes of the graph is kept unchanged, downsampling and upsampling operations are not used. As a result, while the method in~\citep{rong2019dropedge} is easy to integrate to various GNN architectures, the regularization effect is not as strong as the proposed method, which alters both edges and nodes. 
{\color{black}
Closest in spirit to our work is GRAND~\citep{feng2020graph}, where the authors use the Bernoulli distribution to randomly replace node feature vectors with zero vectors. Still, GRAND keeps the structure of the original graph unchanged while we modify the original graph's structure, changing node transition probabilities. Besides, GRAND leverages multiple branches to augment the graph features while we rely on a single branch, which is simpler.  Finally, different from GRAND, we show that the proposed DropNode can be applied to fully supervised tasks such as graph classification.
}

\section{Preliminaries}
\label{sec:preliminaries}
\subsection{Notation and Problem Formulation}
Let $G = (V, E)$ denote an undirected graph, where $V$ is the set of nodes and $E$ is the set of edges. $\mathbf{A} \in \mathbb{R}^{N \times N}$ is the adjacency matrix of $G$, where $\mathbf{A}_{ij}$ is the weight of the corresponding edge connecting node $i$ and node $j$. As $G$ is an undirected graph, the adjacency matrix $\mathbf{A}$ is symmetric (\ie $\mathbf{A}_{ij} = \mathbf{A}_{ji}$). 
The diagonal matrix $\mathbf{D}$  (where $\mathbf{D}_{ii} = \sum_{j=1}^N\mathbf{A}_{ij}$) denotes the degree matrix of $G$. 
The unnormalized Laplacian matrix of $G$ is given by $\mathbf{L} = \mathbf{D} - \mathbf{A}$~\citep{ortega2018graph}. 
The graph $G$ can be fully represented using either the Laplacian or the adjacency matrix. 
Each node of $G$ is characterized by a feature vector $\mathbf{x} \in \mathbb{R}^F$, where $F$ is the feature dimensionality. 
Thus, all the features of the nodes in $G$ can be represented in the form of a matrix, $\mathbf{X} \in \mathbb{R}^{N \times F}$ where each row corresponds to a node and each column corresponds to a feature.

Two common tasks that graph neural networks often address are (i) node and (ii) graph classification. Task (i) concerns predicting a label for each node in a given graph. Specifically, for a graph $G = (V, E)$, the set of labelled nodes can be symbolized as $V_L$  and the set of unlabelled nodes can be symbolized as $V_U$, such that $V = V_L \cup V_U$.
The goal of task (i) is to predict the labels for the nodes in $V_U$. 
In the case that the features and the connections of these nodes are available during the training phase, the task follows the so-called~\textit{transductive setting} ~\citep{kipf2016semi,velivckovic2017graph}. 
On the other hand, in the case that neither the features nor the connections of nodes in $V_U$ are given during training, the task follows the~\textit{inductive setting}~\citep{hamilton2017inductive,chen2018fastgcn}.  
Different from the node classification task (\ie task (i)), the task (ii) aims at predicting \textit{a label for each graph}. For this task, only the inductive setting is considered, that is, given a training set of graphs $S_T = \{ G_1, G_2, \dots, G_M \}$ and the corresponding set of labels $L_T = \{ 1, 2, \dots, C \}^M$ where $C$ is the number of classes and $M$ is the number of examples in the training set, the goal is to learn a prediction function that maps an input graph to a label in $L_T$. 
In this work, we will consider both the node and graph classification tasks. 

\subsection{Random Walk and Node Transition Matrix}
\label{sec:randomwalk_transition}
Random Walk is one of the most effective methods to explore a graph (see~\cite{leskovec2006sampling}). Given a graph $G = (V, E)$,  a length constant $l$, and a starting node $u_0 \in V$, this method generates a ``walk'' of length $l$, represented by a sequence of nodes $\{u_0,u_1,\dots,u_l\}$. A node is sampled for step $k$ in the walk, $k \in [1,l]$, following a distribution $P(u_k | u_{k-1})$, which is expressed by:
\begin{equation}
    \label{eq:random_walk}
    P(u_k = v_j| u_{k-1} = v_i) = \begin{cases} 
        P_{ij},~\text{if}~(v_i, v_j) \in E \\
        0,~\text{if}~(v_i, v_j) \notin E.
    \end{cases}
\end{equation}

In~\equref{eq:random_walk}, $P_{ij}$ is the probability of transitioning from node $v_i$ to node $v_j$. As such, $P_{ij}$ is often referred to as the \textit{node transition probability}. The standard way to compute the transition probabilities is by dividing the weight of the edge $(v_i, v_j)$ by the degree of node $v_i$ (\ie $P_{ij} = \frac{\mathbf{A}_{ij}}{\mathbf{D}_{ii}}$). In general, $P_{ij} \ne P_{ji}$ since the degree of node $v_i$ is not equal to the degree of node $v_j$. 
All the transition probabilities of the nodes in $G$ can be represented in the form of a matrix $\mathbf{P} \in \mathbb{R}^{N \times N}$, where:
\begin{equation}
    \mathbf{P} = \mathbf{D}^{-1} \mathbf{A}.
\end{equation}
Note that $\sum_{j=1}^N \mathbf{P}_{ij} = 1,~\forall i \in [1, N]$.
\section{The Proposed Models}
\label{sec:method}
In this section, we present in detail the contribution of this work. At the beginning, we describe existing message passing schemes~\citep{kipf2016semi,zhang2018}. Afterwards, we introduce the proposed scheme (which follows node transition-based approaches) that leads to a new graph convolutional layer formulation. Subsequently, we present DropNode, a simple yet effective generic regularization method, and we compare it with existing methods that are often employed to regularize GCNN models.
\subsection{Graph Convolutional Layers}
\label{sec:graph_conv_layers}
\subsubsection{Existing Formulations}
\label{sec:existing_formulation}
As presented earlier in the Introduction section, the formulation of a graph convolutional layer in a GCNN model fully expresses the underlying message passing scheme.  Various graph convolutional layers have been proposed in the literature that generally follow the same formulation:
\begin{equation}
    \label{eq:conv_generic}
    \mathbf{H}^{(l+1)} = \sigma (\mathbf{M} \mathbf{H}^{(l)} \mathbf{W}),
\end{equation}
where $\mathbf{H}^{(l)} \in \mathbb{R}^{N \times F}$ is the input to the $l$-th layer, $\mathbf{W} \in \mathbb{R}^{F \times K}$ is the weight matrix of this layer, $\mathbf{M} \in \mathbb{R}^{N \times N}$ is a matrix containing the \textit{aggregation coefficients}, and $\sigma$ is a non-linear activation function such as the sigmoid or the ReLU. 

The message passing scheme, as shown in~\equref{eq:conv_generic}, can be broken down into two substeps, namely the \textit{aggregate} and the \textit{update} substeps. In the former, node features are first transformed by a linear projection, which is parameterized by learnable weights $\mathbf{W}$. Subsequently, for a node in the graph, the features of its neighbors are aggregated by performing a weighted sum with the corresponding weights defined in $\mathbf{M}$. This substep can be seen as a \textit{message} aggregation stage from the neighbors of each node. In the \textit{update} substep, a non-linear transformation is applied to the aggregated features to produce new representations $\mathbf{H}^{(l+1)}$ for the nodes. 

The aggregation coefficient matrix $\mathbf{M}$ defines the message passing scheme, \ie the way in which features are exchanged between nodes. 
An entry $\mathbf{M}_{ij}$ of $\mathbf{M}$ represents the coefficient assigned to features of a source node $j$ when being aggregated toward a destination node $i$. 
As such, the $i$-th row of $\mathbf{M}$ specifies the weights assigned to features of all the nodes when being aggregated toward node $i$. The $j-$th column of $\mathbf{M}$ specifies the weights assigned to a node $j$ when being aggregated toward all the other nodes. 
Intuitively, the rows and the columns of matrix $\mathbf{M}$ determine the influence of the nodes on a destination node, and the influence of a source node on the other nodes, respectively. 
By defining the aggregation coefficient matrix $\mathbf{M}$, one can specify a message passing scheme. 
For instance, in a GCN model~\citep{kipf2016semi}, $\mathbf{M}$ is computed by:
\begin{equation}
    \label{eq:gcn}
    \mathbf{M} = \mathbf{\tilde{D}}^{-\frac{1}{2}} \mathbf{\tilde{A}} \mathbf{\tilde{D}}^{-\frac{1}{2}},
\end{equation}
with $\mathbf{\tilde{A}} = \mathbf{A} + \mathbf{I}_N$ and $\mathbf{I}_N \in \mathbb{R}^{N \times N}$ is an identity matrix. 
By adding the identity matrix $\mathbf{I}_N$, self-connections are taken into account, \ie a node aggregates features from both its neighbors and itself. $\mathbf{\tilde{D}}$ is a diagonal matrix with $\mathbf{\tilde{D}}_{ii} = \sum_{j=1}^{N}\mathbf{\tilde{A}}_{ij}$. In~\equref{eq:gcn} each element of $\mathbf{\tilde{A}}$ is normalized by a factor equal to the square root of the product of the degrees of the two corresponding nodes, namely: 
{\color{black}
\begin{equation}
    \label{eq:m_ij_gcn}
    \mathbf{M}_{ij} = \frac{\mathbf{\tilde{A}}_{ij}}{\sqrt{\mathbf{\tilde{D}}_{ii}\mathbf{\tilde{D}}}_{jj}}. 
\end{equation}
}

Unlike GCN, the DGCNN model~\citep{zhang2018} calculates the aggregation coefficients ($\mathbf{M}_{ij}$) as one-hop node transition probabilities ($\mathbf{\tilde{P}}_{ij}$) (see Section~\ref{sec:randomwalk_transition}):
\begin{equation}
    \label{eq:dgcnn_p}
    \mathbf{M} = \mathbf{\tilde{P}} = \mathbf{\tilde{D}}^{-1} \mathbf{\tilde{A}}.
\end{equation}
{\color{black}
Therefore, it can be seen that:
\begin{equation}
    \label{eq:m_ij_dgcnn}
    \mathbf{M}_{ij} = \frac{\mathbf{\tilde{A}}_{ij}}{\mathbf{\tilde{D}}_{ii}}.
\end{equation}
}

Note that in the DGCNN model, the aggregation coefficients are created by \textit{ normalizing the adjacency matrix using the degrees of the destination nodes}. 
Therefore, the neighboring nodes of a destination node have the same influence (\ie weights) on the destination node, even though their popularity level (\eg node degrees) may vary significantly. This is not desired as popular nodes are often connected with many other nodes, thus messages from the popular nodes are not as valuable as messages from nodes with lower level of popularity. 
The issue of popular nodes has been mentioned in~\citep{do2017multiview,rahimi2015twitter}, where connections to these nodes are explicitly removed. In addition, we observe that in many text retrieval weighing schemes such as TF-IDF, the popular terms across documents are given smaller weights compared to less popular terms. This observation leads us to a new message passing formula, which we describe thoroughly in the next section.

\subsubsection{The Proposed Graph Convolutional Layer}
\label{sec:proposed_gcl}
In this work, we propose a message passing scheme that makes use of the node transition probabilities, as in the DGCNN model~\citep{zhang2018}. 
Unlike the DGCNN, we use the degrees  \textit{of the source nodes} instead of the \textit{destination nodes} for the normalization of the aggregation coefficients. 
Specifically, in our scheme, the aggregation coefficient matrix $\mathbf{M}$ is calculated as:

{\color{black}
\begin{align}
    \label{eq:message_passing_matrix}
    \mathbf{M} &= \mathbf{\tilde{A}}^{T} \mathbf{\tilde{D}}^{-1},\\
    \label{eq:m_ij_pgcn}
    \mathbf{M}_{ij} &= \frac{\mathbf{\tilde{A}}_{ji}}{\mathbf{\tilde{D}}_{jj}} .
\end{align}
}

In \equref{eq:message_passing_matrix}, $\mathbf{M}$ contains one-hop node transition probabilities.  $\mathbf{M}_{ij}$ is the probability of transitioning from a source node $j$ to a destination node $i$. As these probabilities are normalized using the degrees of the source nodes, we have $\sum_{i=1}^N \mathbf{M}_{ij} = 1, \forall j \in \{1,\dots,N\}$. 
It should be noted that in this case, $\sum_{j=1}^N \mathbf{M}_{ij} \ne 1$, which is different from the DGCNN model (as shown in~\equref{eq:dgcnn_p}). 
{\color{black}
It should be noted that $\mathbf{\tilde{D}}$ is a diagonal matrix, and all the entries in the main diagonal are greater than $0$, thus the inverse matrix of $\mathbf{\tilde{D}}$ can be found by simply calculating the inverse of the individual entries on the main diagonal.
}

Substituting $\mathbf{M}$ in~\equref{eq:message_passing_matrix} into the generic formulation in~\equref{eq:conv_generic}, we obtain the formulation for our graph convolutional layer as follows:
\begin{equation}
    \label{eq:message_passing_new}
    \mathbf{H}^{(l+1)} = \sigma ( \mathbf{\tilde{A}}^T \mathbf{\tilde{D}}^{-1} \mathbf{H}^{(l)} \mathbf{W}).
\end{equation}
We refer to this graph convolutional layer as the transition \textbf{P}robability based \textbf{G}raph \textbf{CONV}olutional layer, abbreviated as GPCONV (for ease of reference, we also use GCONV to refer to the graph convolutional layer proposed by~\citep{kipf2016semi}). 
Intuitively, as the adjacency matrix is normalized by the degrees of source nodes, a node always has the same influence on its neighboring nodes. Specifically, a node's message (a.k.a., feature vector) is disseminated to its neighbors with the same weight. 
        
As a result, a popular node (i.e., high degree) will have a smaller influence on its neighbors. {\color{black}
Considering two nodes, i.e., $v_i$ and $v_j$ with different degrees (i.e., $\text{deg}(v_j) \gg \text{deg}(v_i)$), we can see that:
\begin{equation}
    \mathbf{M}_{ij}(\text{PGCN}) = \frac{\mathbf{\tilde{A}}_{ji}}{\mathbf{\tilde{D}}_{jj}} < \mathbf{M}_{ij}(\text{GCN}) = \frac{\mathbf{\tilde{A}}_{ij}}{\sqrt{\mathbf{\tilde{D}}_{ii}\mathbf{\tilde{D}}}_{jj}} < \mathbf{M}_{ij}(\text{DGCNN}) = \frac{\mathbf{\tilde{A}}_{ij}}{\mathbf{\tilde{D}}_{ii}}.
\end{equation}

Denote by $\mathbf{\tilde{H}}^{(l+1)}$ the result of the aggregation step,~\ie the product of the aggregation coefficient matrix and the input representations, $\mathbf{\tilde{H}}^{(l+1)} = \mathbf{M} \mathbf{H}^{(l)}$. Thus: 
\begin{equation}
    \label{eq:message_passing_explain}
    \mathbf{\tilde{H}}_i^{(l+1)} = \mathbf{M}_{i1} \mathbf{H}_1^{(l)} + \mathbf{M}_{i2} \mathbf{H}_2^{(l)} + \dots + \mathbf{M}_{iN} \mathbf{H}_N^{(l)} \nonumber = \sum_{j=1}^N \mathbf{M}_{ij} \mathbf{H}_{j}^{(l)} ,
\end{equation}
$\forall i \in \{1, \dots, N\}$. Here, $\mathbf{H}_j^{(l)}$ and $\mathbf{\tilde{H}}_i^{(l+1)}$ are row vectors; $\mathbf{H}_j^{(l)}$ is the representation of node $v_j$ at the $l$-th layer. 
It can be seen that the proposed message passing leads to the smallest influence of node $v_j$ on the representation of node $v_i$. 
This behaviour, which we call \textit{popularity penalizing}, is desired
in the case that the nodes in the neighborhood of the popular node (\ie $v_j$) belong to different clusters (\ie these nodes belong to different classes) and these clusters have similar sizes. 
In that case, 
the message from the popular node is not helpful in creating discriminative representations for the neighboring nodes. 

Let us define a coefficient to quantify the \textit{diversity and balance of classes} over the neighborhoods of popular nodes. Denote by $\mathcal{N}_{v_j}$ the set of neighboring nodes of node $v_j$ and let $C(\mathcal{N}_{v_j})$ be the set of the classes of these nodes, namely that $C(\mathcal{N}_{v_j})$ contains distinct classes. The coefficient, denoted by $\gamma$, is defined by:
\begin{equation}
    \label{eq:dispersion}
    \gamma = \frac{4}{|V|} \sum_{i=1}^{|V_h|} \frac{|C(\mathcal{N}_{v_j})|}{|\mathcal{N}_{v_j}|} \cdot \frac{-\sum_{k=1}^{|C(\mathcal{N}_{v_j})|} \frac{c_k}{|\mathcal{N}_{v_j}|} \text{log}(\frac{c_k}{|\mathcal{N}_{v_j}|})}{\text{log} (|C(\mathcal{N}_{v_j})|)},
\end{equation}
where $V_h \subset V$; $V_h$ contains nodes with a degree greater than the $75$-th percentile of all node degrees as we are interested in popular nodes. 
$c_k$ indicates the number of nodes of class $k$. We can see that $\gamma$ is the sum of multiple elements where each element consists of two terms.  The first term represents the diversity of classes over the neighborhood $\mathcal{N}_{v_j}$, ranging in $(0,1)$. 
The second term is the Shannon index~\citep{shannon_diversity}, indicating the balance of the classes in $\mathcal{N}_{v_j}$ and its value is also in the range $(0, 1)$. 
Clearly, $\gamma \in (0, 1)$ and a larger value of $\gamma$ will lead to a better performance of the proposed message passing.
}

\subsubsection{Comparison with Existing Formulations}
%
\begin{figure}[t!]
        \centering
        \includegraphics[scale=0.3]{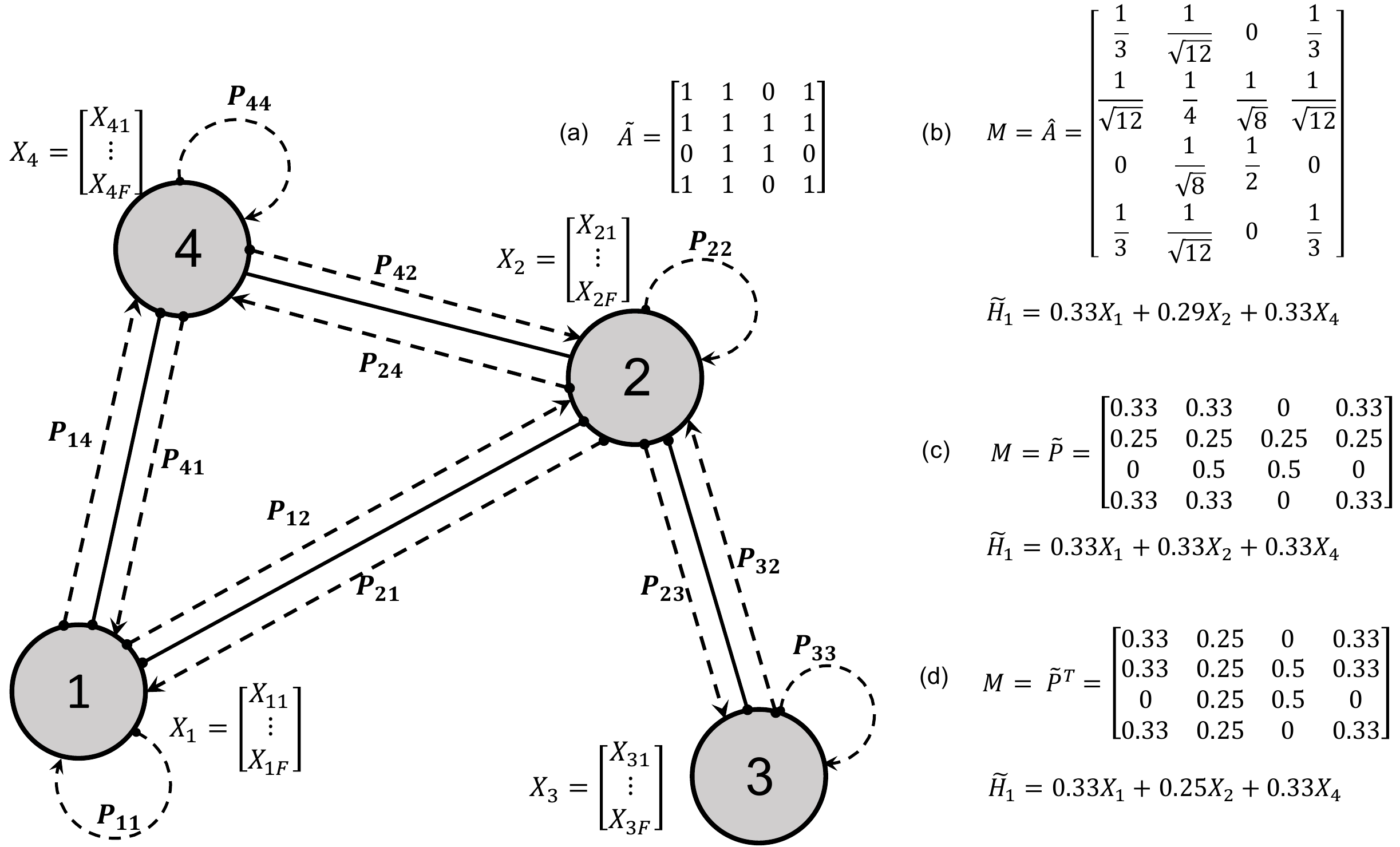}\\
        \caption{
        A graph with four nodes and four edges (solid lines) with directed transition probabilities indicated by dashed lines. 
        (a) shows the adjacency matrix with self-connections added. 
        (b), (c) and (d) show the equations to calculate the aggregated feature representation (\ie $\mathbf{\tilde{H}}_2$) for node $1$, following the formulations in the GCN model~\citep{kipf2016semi}, the DGCNN model~\citep{zhang2018} and the proposed scheme, respectively. 
       We observe that in the proposed scheme, (\textit{i}) the highest-degree node (\ie node 2) is \textit{less influential} in calculating $\mathbf{\tilde{H}}_1$ than the other nodes, which are of lower degrees, and (\textit{ii}) the overall influence of a node on their neighboring nodes are the same (\ie entries in a column, where there exist corresponding connections, have the same value.).}
        \label{fig:simple_graph}
\end{figure}

To illustrate the difference between the proposed graph convolutional layer's formulation and the DGCNN model~\citep{zhang2018}, 
we consider a simple graph containing four nodes as shown in Fig.~\ref{fig:simple_graph}. These nodes are associated with feature vectors $\mathbf{X}_1$, $\mathbf{X}_2$, $\mathbf{X}_3$ and $\mathbf{X}_4 \in \mathbb{R}^F$. We denote with $\mathbf{\tilde{P}} \in \mathbb{R}^{4 \times 4}$ the transition probability matrix, as calculated in~\equref{eq:dgcnn_p}, and the \textit{aggregate} step in the DGCNN model produces an aggregated feature representation for node $1$ as:
\begin{align}
    \label{eq:ex_mp_zhang}
    \begin{split}
    \mathbf{\tilde{H}}_1 &= \mathbf{\tilde{P}}_{11} \mathbf{X}_1 + \mathbf{\tilde{P}}_{12} \mathbf{X}_2 + \mathbf{\tilde{P}}_{13} \mathbf{X}_3 + \mathbf{\tilde{P}}_{14}\mathbf{X}_4\\
    &= 0.33\mathbf{X}_1 + 0.33\mathbf{X}_2 + 0.33\mathbf{X}_4
    \end{split}
\end{align}
Using the proposed scheme (\ie \equref{eq:message_passing_matrix}), we obtain:
\begin{align}
    \label{eq:ex_mp_ti}
    \begin{split}
    \mathbf{\tilde{H}}_1 &= \mathbf{\tilde{P}}_{11} \mathbf{X}_1 + \mathbf{\tilde{P}}_{21} \mathbf{X}_2 + \mathbf{\tilde{P}}_{31} \mathbf{X}_3 + \mathbf{\tilde{P}}_{41}\mathbf{X}_4\\
    &= 0.33\mathbf{X}_1 + 0.25\mathbf{X}_2 + 0.33\mathbf{X}_4
    \end{split}
\end{align}
%

Even though~\equref{eq:ex_mp_ti} has the same form as~\equref{eq:ex_mp_zhang}, the intuitions behind the two {\color{black}are different. 
On one hand, the messages in the DGCNN model are passed following the \textit{transition probabilities from destination nodes to source node}. 
For instance, the message of node $2$ ($\mathbf{X}_2$) is passed with a weight equal to $\mathbf{\tilde{P}}_{12}$, which is the transition probability from node $1$ (destination node) to node $2$ (source node). 
On the other hand, in our formulation, the messages are passed following the \textit{ transition probabilities in the reverse direction}, namely, from the source nodes to destination node. 
This difference results in an important distinction in the way the influence (\ie weights) of the nodes are determined. }
As shown in Fig.\ref{fig:simple_graph}, the DGCNN model gives a weight of $0.33$ to $\mathbf{X}_2$ in calculating $\mathbf{\tilde{H}}_1$ while the proposed scheme uses a weight of $0.25$. Bear in mind that node $2$ has the highest degree of $4$ (self-loop added), we can conclude that the proposed scheme gives lower weight to nodes with a higher degree. 
This leads to an effect similar to the TF-IDF weighting scheme as discussed in Section~\ref{sec:existing_formulation}. 
Next, we will present how we build GCNN models, with the GPCONV layer as the main building block, for node and graph classification tasks.


\subsection{Node and Graph Classification Models}
\label{sec:basic_model}
Using the proposed GPCONV layer as a building block, we construct GCNN models for the \textit{node} and \textit{graph classification} tasks. 
\subsubsection{Node Classification Model}
\label{sec:node_classification_model}
Our node classification model consists of $L$ GPCONV layers. The model's operation can be expressed by: 
\begin{equation}
    \label{eq:node_classification_model}
    \mathbf{O} = \text{softmax}\big( \mathbf{M} \sigma \big( \dots \sigma \big( \mathbf{M} \mathbf{X} \mathbf{W}_1 \big) \dots \big) \mathbf{W}_L \big). 
\end{equation}
In this \equref{eq:node_classification_model}, $\mathbf{M}$ contains transition probabilities calculated using~\equref{eq:message_passing_matrix}, $\mathbf{O} \in \mathbb{R}^{N \times C}$ are the predicted class probabilities where $C$ is the number of classes. Note that the number of GPCONV layers, $L_{\text{GPCONV}}$, is a design choice. 
We refer to this model as the $\text{PGCN}_n$ model where P refers to transition probabilities and $n$ stands for node classification.
\subsubsection{Graph Classification Model}
\label{sec:graph_classification_model}
For graph the classification task, we need to predict a single label for the whole graph. 
To address this task, we design a model (similar to the node classification one) with a global pooling and several fully-connected layers added on top of the last GPCONV layer. 
The pooling operator is employed to produce a single representation for the whole graph, while the fully-connected layers and a softmax classifier are used to output the predicted label probabilities. 
There are several pooling techniques for graph classification, such as \texttt{SortPooling}~\citep{zhang2018}, \texttt{DiffPool}~\citep{ying2018hierarchical}, \texttt{Top-K-Selection}~\citep{gao2019graph}, \texttt{max-pooling}~\citep{zhang2018} and \texttt{mean-pooling}~\citep{simonovsky2017dynamic,monti2019fake,bianchi2019graph}. 
In our model, we use the global \texttt{mean-pooling} as it has been proven effective and is widely used for the task of graph classification~\citep{monti2019fake,bianchi2019graph}. 
We refer to our graph classification model as the $\text{PGCN}_g$. Here, $g$ stands for graph classification, which differentiates it from the proposed node classification model $\text{PGCN}_n$. The $\text{PGCN}_g$ can be expressed by:
\begin{align}
    \label{eq:propagation_rule}
    \mathbf{H} &=   \sigma \big( \mathbf{M} \sigma \big( \dots \sigma \big( \mathbf{M} \mathbf{X} \mathbf{W}_1 \big) \dots \big) \mathbf{W}_L \big)\\
    \label{eq:mean_pooling}
    \mathbf{h} &= \texttt{mean-pooling} \big( \mathbf{H} \big)\\
    \label{eq:fc}
    \mathbf{o} &= \texttt{softmax} \big( \text{FC} \big( \mathbf{h} \big)  \big) 
\end{align}

The~\equref{eq:propagation_rule},~\eqref{eq:mean_pooling},~\eqref{eq:fc} are written for one graph, so $\mathbf{H} \in \mathbb{R}^{N \times K}$ is a $2-$D matrix, $\mathbf{h} \in \mathbb{R}^K$ and $\mathbf{o} \in \mathbb{R}^C$ are vectors. 
FC indicates the fully-connection part of the $\text{PGCN}_g$ model. 
In practice, multiple graphs can be stacked together by concatenating the corresponding adjacency matrices diagonally and concatenating the feature matrices vertically. In that case, the pooling operation in~\equref{eq:mean_pooling} is applied graph-wise.
\subsection{DropNode Regularizer}
In this section, we present the proposed DropNode regularization method in detail. 
After that, we show how this method could be further incorporated into the node and graph classification models introduced in Section~\ref{sec:basic_model}, 
and finally we describe the connection between DropNode and the well-known dropout regularization method~\citep{srivastava2014dropout}. 
\subsubsection{DropNode}
\label{sec:dropnode}
%

The basic idea behind DropNode is to randomly sample sub-graphs from an input graph at each training iteration. 
This is achieved by dropping nodes following a Bernoulli distribution with a pre-defined probability $1-p$, $p \in (0,1)$. 
{\color{black}
Furthermore, an edge is retained if its nodes are kept; otherwise, the edge is discarded.}

{\color{black}Such node dropping procedure can be seen as a \textit{downsampling} operation. 
Therefore, DropNode can be implemented as an individual downsampling layer, which can be integrated to different levels of a graph neural network. 
Denote by $\mathbf{H}^{(l)} \in \mathbb{R}^{N \times K_l}$ the input of the $l$-th downsampling layer, the following equations describe how DropNode works:

\begin{align}
    \label{eq:r_index}
    \mathbf{r} &= (r_1, \dots, r_N)^T ,\\
    \label{eq:bermoulli}
    \mathbf{r}_i &\sim \texttt{Bernoulli}(p) , i \in \{1, \dots, N\}, \\
    \label{eq:select_rows}
    \mathbf{H}^{(l+1)}, \mathbf{ID}^{(l)} &= \texttt{SELECT} (\mathbf{H}^{(l)}, \mathbf{r}) .
\end{align}

In~\eqref{eq:r_index}, $\mathbf{r}$ is \giannis{a} vector containing i.i.d random variables responsible for deciding which nodes should be kept. In~\eqref{eq:select_rows}, only rows corresponding with $\mathbf{r}_i = 1$ are retained, leading to a sub-matrix $\mathbf{H}^{(l+1)} \in \mathbb{R}^{N_l \times K_l}$, where $N_l = pN$ on average. 
The second output $\mathbf{ID}^{(l)}$ is a vector containing indices of selected rows, corresponding to the retained nodes. 
The output of the downsampling layer can become the input for a graph convolutional layer. Thus, the adjacency matrix also needs to be modified by
\begin{align}
    \label{eq:new_adj}
    \mathbf{\tilde{A}}^{(l+1)} = \texttt{SELECT}(\mathbf{\tilde{A}}^{(l)}, \mathbf{r}) ,
\end{align}
where only rows and columns corresponding to $\mathbf{r}_i = 1$ are kept. 
The new adjacency matrix in~\eqref{eq:new_adj} is then used for computing the aggregation coefficient matrix $\mathbf{M}$ following the chosen message passing formulation (see Section~\ref{sec:graph_conv_layers}).
}

{\color{black}
Depending on specific tasks (e.g., node classification), \textit{upsampling layers} might be needed to reconstruct the original graph structure.
An upsampling layer is paired with a downsampling layer and uses the index vector of the downsampling layer for the reconstruction. }
Specifically, suppose the $l$-th downsampling layer is paired with the $k$-th  upsampling layer where $k > l$, 
this upsampling layer takes as input a matrix $\mathbf{H}^{(k)} \in \mathbb{R}^{N_l \times K_k}$ and produces an output matrix $\mathbf{H}^{(k+1)} \in \mathbb{R}^{N \times K_k}$. 
Each row in $\mathbf{H}^{(k)}$ is copied to a row in $\mathbf{H}^{(k+1)}$ according to the index vector {\color{black} $\mathbf{ID}^{(l)}$} obtained by the $l$-th downsamling layer. The rows in $\mathbf{H}^{(k+1)}$ that do not correspond to a row in $\mathbf{H}^{(l)}$ are filled in with zeros. 

Similar to dropout, the proposed DropNode method operates only during the training phase. During the testing phase, all the nodes in the graph are used for prediction. 
We should note that in the case that the upsampling layers are not employed, the output of each downsampling layer needs to be scaled by a factor of $\frac{1}{p}$. 
This scaling operation is to maintain the same expected outputs for neurons in the subsequent layer during the training and testing phases (similar to dropout).

An important property of DropNode is that one or multiple sub-graphs are randomly sampled at each training iteration. Hence, the model does not see all the nodes during the training phase. As a result, the model should not rely on only a single prominent local pattern or on a small number of nodes, but to leverage information from all the nodes in the graph. The risk for the model to memorize the training samples, therefore, is reduced, avoiding over-fitting. In addition, the model is trained using multiple deformed versions of the original graphs. This can be considered as a data augmentation procedure, which is often used as an effective regularization method~\citep{devries2017improved}. 
On the other hand, the DropNode method reduces connectivity between nodes in the graph. Lower connectivity helps alleviate the smoothing of representation of the nodes when the GCNN model becomes deeper~\citep{rong2019dropedge}. As a result, features of nodes in different clusters will be more distinguishable, which could lead to an improved performance on different downstream tasks for deep GCNN models. 

\subsubsection{Dropping Strategies}
{\color{black}
In~\secref{sec:dropnode}, we have presented DropNode, which relies on the Bernoulli distribution to sample sub-graphs from the original graph. While this strategy is simple, it is structure-agnostic. In literature, there are many approaches for sampling graphs, including Random Page Rank Node (RPN), Random Edge (RE), Random Node Neighbo (RNN) and Random Walk (RW)~\citep{leskovec2006sampling}. It is found that RW is one of the most effective graph sampling approaches w.r.t. preserving graph properties such as clustering coefficient~\citep{leskovec2006sampling}. 
Therefore, it is tempting to investigate the RW-based node dropping for DropNode. 
In this section, we look into the detailed effect of the Bernoulli-based dropping introduced earlier and the RW-based dropping.

Let us first study the Bernoulli-based dropping approach (B-dropping) by considering a graph $G = (V, E)$, where $|V| = N$ and $|E| =  M$. Thus the probability of the existence of an edge of $G$ is $p_{(u,v)} = \frac{M}{{\binom{N}{2}}}$.
As the probability for selecting a node is $p$, on average $n = \floor{pN}$ nodes will be retained. Sub-graphs thus have the expected number of edges as follows:
\begin{equation}
    \label{eq:expected_edges}
    \overline{m} = \frac{M \binom{n}{2}}{\binom{N}{2}}.
\end{equation}
Therefore, the sub-graphs created by the B-dropping approach can be modelled by an Erd\H{o}s–R\'{e}nyi (ER) process~\citep{erdHos1960evolution}, with $n = \floor{pN}$ nodes and $\overline{m}$ edges. We will show later that the sub-graphs created by the B-dropping have some properties of the ER model.

Different from the B-dropping, the RW-dropping statistically maintains the local structure of the underlying graph. The RW-dropping follows the random walk procedure introduced in~\secref{sec:randomwalk_transition} with two small modifications. First, if a node is re-visited, we ignore it and move to another node. Furthermore, if the walk is stuck (e.g., if we visit a small connected component), we jump to a node which has not been visited. 
These modifications will allow a deterministic number of nodes for each sampling step. We will see later that the RW-dropping approach tends to preserve local clusters, leading to a  clustering coefficient similar to that of the original graph. Eventually, this approach tends to produce small-world sub-graphs~\citep{watts1998collective}. 
As a result, RW-dropping may result in better node representations, leading to better performance for downstream tasks. However, as the RW-dropping has to be done step by step, thus this approach is slower compared to the B-dropping.
}

\subsubsection{Incorporating DropNode into the \texorpdfstring{$\text{PGCN}_n$}{2} Model}
\label{sec:pgcn_n}

\begin{figure*}[t!]
    \centering
    \includegraphics[width=\linewidth]{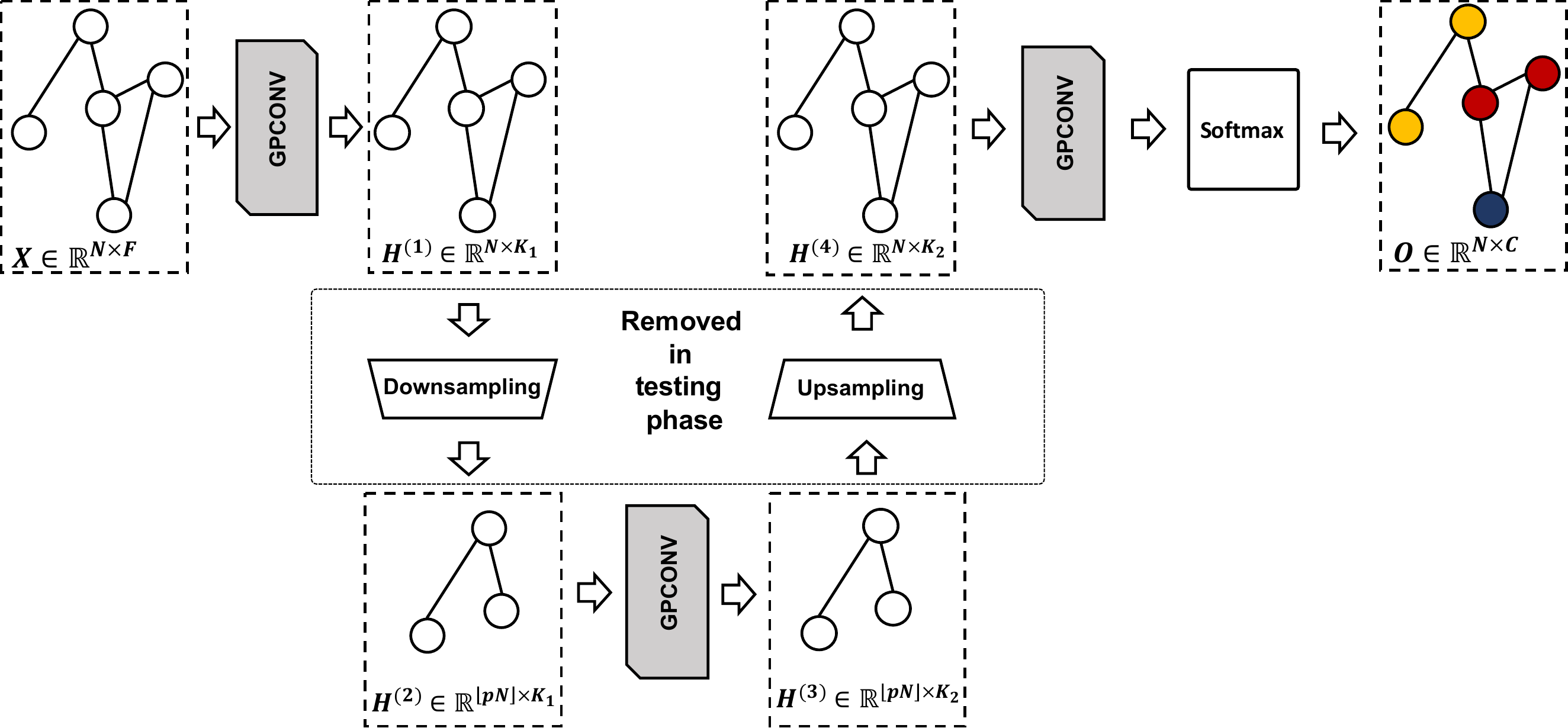}\\
    \caption{The structure of the $\text{PGCN}_n + \text{DropNode}$ model with two GPCONV layers and one pair of downsampling-upsampling layers for the node classification task.}
    \label{fig:node_classification}
\end{figure*}
\Figref{fig:node_classification} shows an example of how we incorporate DropNode into a $\text{PGCN}_n$ model for the node classification task. 
In this example, we employ three GPCONV layers and a pair of downsampling and upsampling layers. 
Given a graph $G$ with input features $\mathbf{X}$, the model transforms $\mathbf{X}$ through the first GPCONV layer into $\mathbf{H}^{(1)}$. 
Subsequently, the downsampling layer randomly drops a subset of rows of the transformed feature vectors $\mathbf{H}^{(1)}$. The remaining features $\mathbf{H}^{(2)}$ are fed into the second GPCONV layer. 
For the node classification task, it is essential to keep the same number of nodes at the output. 
To this end, an upsampling layer {\color{black}(see~\secref{sec:dropnode})} is used to reconstruct the original graph structure. The output of this layer serves as input to the third GPCONV and a softmax classifier to produce 
class probabilities for all the nodes of the graph. 
It should be noted that the number of GPCONV layers $L_{\text{GPCONV}}$ and downsampling-upsampling layer pairs $L_{\text{DU}}$ are hyper-parameters of this model, {\color{black}thus more complex architectures can be formed by increasing these parameters. }
We refer to this model as the $\text{PGCN}_n + \text{DropNode}$ model.

\subsubsection{Incorporating DropNode into the \texorpdfstring{$\text{PGCN}_g$}{2} Model}
\begin{figure*}[t!]
    \centering
    \includegraphics[width=\linewidth]{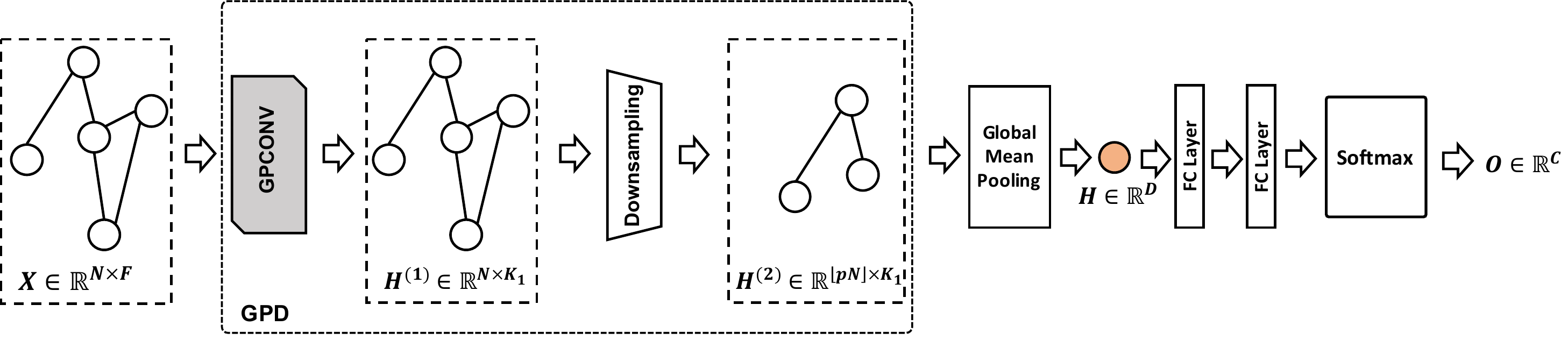}\\
    \caption{The $\text{PGCN}_g + \text{DropNode}$ model for graph classification. A GPCONV layer is combined with a downsampling layer to create a GPD block. Several GPD blocks can be stacked together to achieve better expressiveness power. 
    The output of the final GPD block will be globally mean pooled making the final representation of the graph as denoted by $\mathbf{H}$. One or several fully-connected layer(s) (FC layers) and a softmax classifier are employed to predict the label of the graph.}
    \label{fig:graph_classification}
\end{figure*}
\Figref{fig:graph_classification} shows an example of how DropNode can be integrated with the $\text{PGCN}_g$ model for the graph classification task. We refer to this model as $\text{PGCN}_g + \text{DropNode}$. 
The model has a block consisting of one GPCONV layer and one downsampling layer, which is abbreviated to ``GPD'' block. 
The GPD block is followed by a simple mean-pooling layer which produces a single representation vector for the whole graph. 
{\color{black}
While there exist several more advanced pooling techniques such as DiffPool~\citep{ying2018hierarchical} and $k$-max pooling~\citep{gao2019graph}, we choose the mean-pooling as it is simple and  pooling is not a focus of this work.
}
Then, two fully-connected layers act as a classifier on that representation vector. Similar to the $\text{PGCN}_n + \text{DropNode}$ model (see \secref{sec:pgcn_n}), the numbers of GPD blocks $L_{\text{GPD}}$ and fully-connected layers $L_{\text{FC}}$ are hyper-parameters of the $\text{PGCN}_g + \text{DropNode}$ model. 
In graph classification, the reconstruction of the graph structure is not needed {\color{black}as we only need a single label for an input graph} (\ie unlike the case of the node classification task). As such, it is not necessary to employ upsampling layers in the $\text{PGCN}_g + \text{DropNode}$ model. 
As a result, the outputs of each downsampling layer need to be scaled by a factor of $\frac{1}{p}$ during training as mentioned earlier in~\secref{sec:dropnode}. 
\subsubsection{Connection between DropNode and Dropout}
\begin{figure}[t!]
        \centering
        \includegraphics[width=0.7\linewidth]{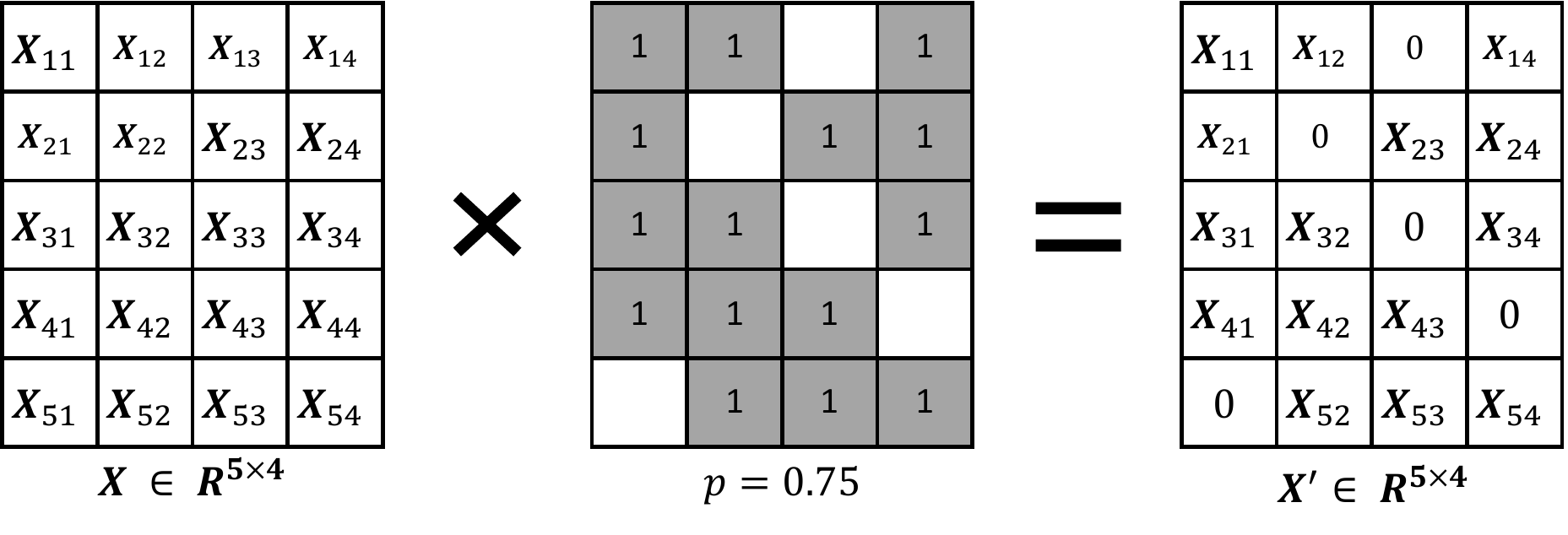}\\
        \caption{Dropout applied to feature matrix $\mathbf{X}$. Several neural units corresponding to elements on each row of $\mathbf{X}$ are randomly deactivated during the training phase. 
        This operation produces matrix $\mathbf{X}^{\prime}$ of the same shape as the input.}
        \label{fig:dropout}
\end{figure}

\begin{figure}[t]
        \centering
        \includegraphics[width=0.6\linewidth]{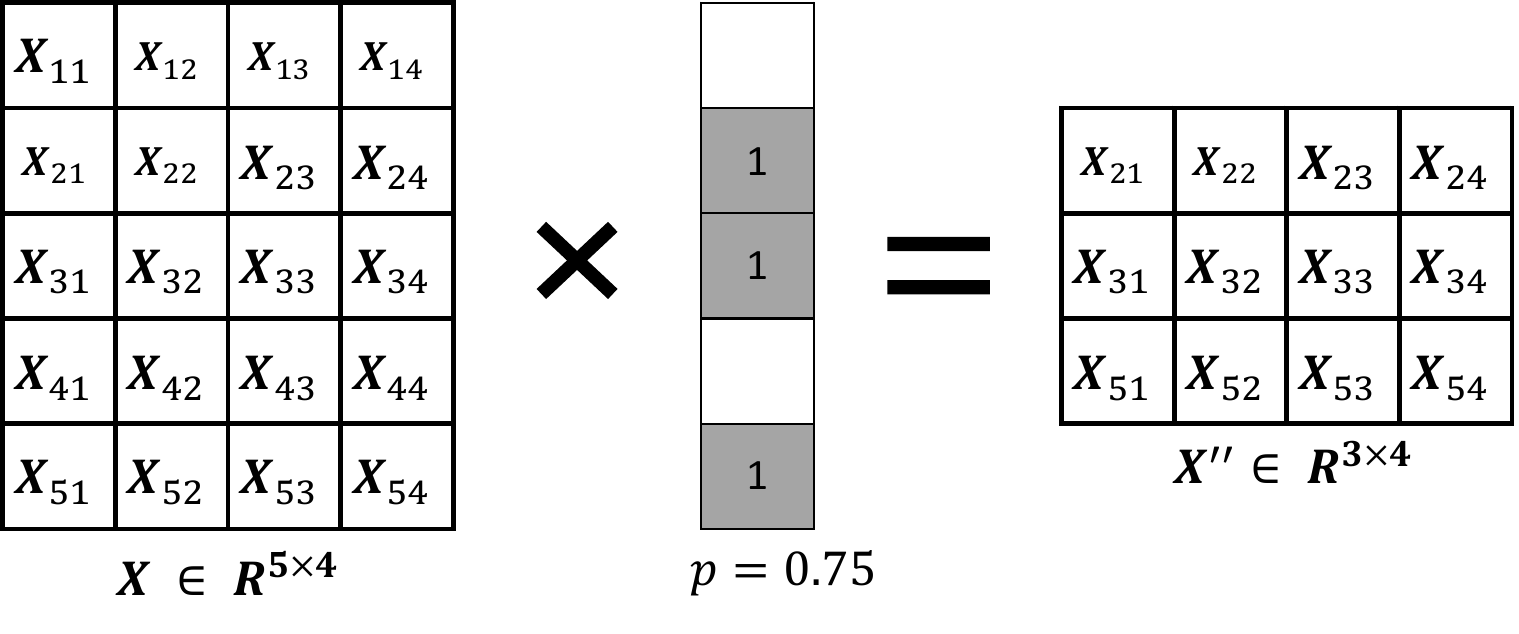}\\
        \caption{DropNode applied to feature matrix $\mathbf{X}$. Unlike dropout, some rows are randomly eliminated, leading to matrix $\mathbf{X}''$ with smaller number of rows.}
        \label{fig:dropnode}
\end{figure}

DropNode and dropout both involve dropping activations in a layer of a neural network, yet, they are conceptually different. 
In DropNode, all the activations of a node are dropped. On the contrary, in dropout, the dropping is distributed across the nodes, so that on average, each node has a ratio of activations dropped. 
In addition, with DropNode, the structure of the graph is altered, while this does not happen in dropout. 
Implementation-wise, let $\mathbf{X} \in \mathbb{R}^{N\times K}$ be the input, 
dropout produces a matrix $\mathbf{X}' \in \mathbb{R}^{N \times K}$ of the same dimensions as $\mathbf{X}$ in which some entries are randomly set to zero; whereas, DropNode produces a sub-matrix $\mathbf{X}'' \in \mathbb{R}^{\floor{pN}\times K}$ where $p \in (0, 1)$. 
Figures~\ref{fig:dropout},~\ref{fig:dropnode} show the difference between dropout and DropNode.

%

\section{Experiments}
\label{sec:experiment}

\subsection{Datasets}
\begin{table}[t]
\centering
\caption{Datasets for the node classification tasks. $+/-$ indicates whether the corresponding features are available or unavailable. The numbers in parentheses denote the dimensionality of the corresponding feature vectors.
{\color{black} $\gamma$ is the diversity-balance coefficient defined in~\secref{sec:proposed_gcl}.
}
}
\label{table:node_classification_dataset}
\footnotesize
 \begin{tabular}{c | c | c | c } 
 \hline \hline
  & CORA & CITESEER & PUBMED \\
\hline
\hline
\# Nodes & $2708$ & $3327$ & $19,717$ \\
\hline
\# Edges & $5429$ & $4732$ & $44,338$ \\
\hline
\# Labels & $7$ & $6$ & $3$ \\
\hline
Node Attr. & $+(1433)$ & $+(3703)$ & $+(500)$ \\
\hline
Edge Attr. & $-$ & $-$ & $-$ \\
\hline 
$\gamma$ & $0.15$ & $0.33$ & $0.13$\\
\hline \hline
\end{tabular}
\end{table}

\begin{table}[t]
\centering
\caption{Datasets for graph classification task. $+/-$ indicates whether the corresponding features are available or unavailable. The numbers in parentheses denote the dimensionality of the corresponding feature vectors. 
{\color{black} $\overline{\gamma}$ is the average diversity-balance coefficient calculated on a set of graphs.
}
}
\label{table:graph_classification_dataset}
\footnotesize
 \begin{tabular}{c | c | c | c | c | c} 
 \hline \hline
  & PROTEINS & D\&D & ENZYMES & MUTAG & NCI1\\
\hline
\hline
\# Graph & $1113$ & $1178$ & $600$ & $188$ & $4110$\\
\hline
\# Label & $2$ & $2$ & $6$ & $2$ & $2$\\
\hline
\# Avg. Node & $39.06$ & $284.32$ & $32.63$ & $17.93$ & $29.87$\\
\hline
\# Avg. Edge & $72.82$ & $715.66$ & $62.14$ & $19.79$ & $32.30$  \\
\hline
Node Label & $+$ & $+$ & $+$ & $+$ & $+$ \\
\hline
Edge Label & $-$ & $-$ & $-$ & $+$ & $-$ \\
\hline
Node Attr. & $+(29)$ & $-$ & $+(18)$ & $-$ & $-$ \\
\hline
Edge Attr. & $-$ & $-$ & $-$ & $-$ & $-$\\
\hline
$\overline{\gamma}$ & $0.25$ & $0.56$ & $0.29$ & $0.0005$ & $0.03$\\
\hline \hline
\end{tabular}
\end{table}

In our experiments, we consider both node and graph classification tasks. 
For node classification, we use three benchmark citation network datasets\footnote{\url{https://linqs.soe.ucsc.edu/data}}, namely, CORA, CITESEER and PUBMED~\citep{sen2008collective}. 
Graphs are created for these datasets by considering scientific papers as nodes and references between the papers as edges. 
Each node is represented by a bag-of-words feature vector extracted from the corresponding document. In these datasets, all the nodes are labeled. 
Following Kipf~\etal~\citep{kipf2016semi} and Veli{\v{c}}kovi{\'c}~\etal~\citep{velivckovic2017graph}, we consider the graphs of CORA, CITESEER and PUBMED  \textit{undirected}, although the references are actually directed. 
The description of the considered datasets for node classification is presented in~\tabref{table:node_classification_dataset}.

Concerning the graph classification task, we employ the following datasets\footnote{\url{https://ls11-www.cs.tu-dortmund.de/staff/morris/graphkerneldatasets}}: the bioinformatics datasets, namely ENZYMES, PROTEINS, D\&D, MUTAG; the scientific collaboration dataset COLLAB~\citep{KKMMN2016}; the chemical compound dataset NCI1. 
In the bioinformatics datasets, each graph represents a biological structure. 
For the COLLAB dataset, a graph represents an ego-network of researchers who have collaborated with each other~\citep{gomez2017dynamics}. 
The NCI1 dataset represents the activity against non-small cell lung cancer~\citep{gomez2017dynamics}. 
The description of these datasets is presented in \tabref{table:graph_classification_dataset}.
\subsection{Experimental Setup}
For both node and graph classification tasks, we employ classification accuracy as the performance metric. 
Concerning the node classification task, similar to~\citep{kipf2016semi,velivckovic2017graph}, we consider the \textit{transductive} experimental setting (see Section~\ref{sec:preliminaries}). We employ the standard train/validation/test set splits of all the considered datasets for node classification to guarantee a fair comparison with prior works~\citep{kipf2016semi,velivckovic2017graph,gao2019graph}. That is, $(140/500/1000)$ nodes are used for the CORA dataset, $(120/500/1000)$ nodes are used for the CITESEER dataset and $(60/500/1000)$ nodes are used for the PUBMED dataset. 
It is worth mentioning that the amount of labelled nodes is much smaller than the amount of test nodes, which makes the task highly challenging. 
Similar to~\citep{velivckovic2017graph,gao2019graph,velivckovic2018deep}, we report the mean and standard deviation of the results over $100$ runs with random weight initialization. 

Concerning the graph classification task, following existing works~\citep{ying2018hierarchical,zhang2018}, we employ a $10$-fold cross validation procedure and report the average accuracy over the folds. Among the graph classification datasets, only PROTEINS and ENZYMES provide node features (see~\tabref{table:graph_classification_dataset}), which can be used directly as input to the proposed models. For the rest of the datasets, we use the degree and the labels of the nodes as features. 

We compare the performance of our models with state-of-the-art baseline models. Specifically, for the node classification task, the selected baselines are the GCN~\citep{kipf2016semi}, GAT~\citep{velivckovic2017graph}, GraphSAGE~\citep{hamilton2017inductive}, DGI~\citep{velivckovic2018deep}, GMNN~\citep{qu2019gmnn}, Graph U-Net~\citep{gao2019graph}, DGCNN~\citep{zhang2018}, and DropEdge~\citep{rong2019dropedge} models. 
The DGCNN model is originally designed for graph classification. In order to use this model for node classification, we employ only its message passing mechanism (see Section \ref{sec:existing_formulation}), ignoring its pooling part. 
For the graph classification task, the Graph U-Net~\citep{gao2019graph}, DGCNN~\citep{zhang2018},  DiffPool~\citep{ying2018hierarchical}, GraphSAGE~\citep{hamilton2017inductive}, CapsGNN~\citep{xinyi2018capsule} and SAGPool~\citep{lee2019self} models are selected. 
In addition, following ~\citep{luzhnica2019graph}, we employ two simple baseslines including a fully-connected neural network with two hidden layers denoted by FCN, and a combination of FCN with one graph convolutional layer (GCONV) from the popular GCN model~\citep{kipf2016semi} (see Section \secref{sec:existing_formulation}) denoted as GCN + 2FC. 
For each baseline model and a benchmark dataset, we select the best results reported in the corresponding paper (if available). Otherwise, we collect the results using either the implementations released by the corresponding authors or self-implemented source code. 

\subsection{Hyperparameter Settings}
\label{sec:hyperparameter_setting}
Hyperparameters of the proposed models are found empirically via tuning. 
For the node classification model $\text{PGCN}_n$, we use two GPCONV layers ($L_{\text{GPCONV}} = 2$). 
This choice also follows the best configuration suggested in~\citep{kipf2016semi}. 
Each GPCONV layer has a hidden dimension of $64$. In addition, dropout is added after each GPCONV layer with a dropping rate of $0.7$. 
When DropNode is used (\ie the $\text{PGCN}_n + \text{DropNode}$ model), we employ three GPCONV layers ($L_{\text{GPCONV}} = 3$), also with a hidden dimension of $64$, and a pair of downsampling-upsampling layers. 
Compared to the $\text{PGCN}_n$ model, one additional GPCONV layer is added between the downsampling and upsampling layers. 
The downsampling layer of DropNode has the keep ratio $p$ selected in such a way that $200$ nodes are retained for all the datasets. We do not employ dropout for $\text{PGCN}_n + \text{DropNode}$ as the DropNode has already had the regularization effect on the considered model. We train the two models with learning rates of $0.01$ and $0.001$, respectively.

For our graph classification models, namely, $\text{PGCN}_g$ and $\text{PGCN}_g + \text{DropNode}$, we employ one GPCONV layer ($L_{\text{GPCONV}} = 1$) and a single GPD block ($L_{\text{GPD}} = 1$), respectively. Both models use two fully-connected layers ($L_{\text{FC} = 2}$). The GPCONV and the fully-connected layers have $512$ hidden units each. We employ dropout after each layer with a dropping ratio of $0.5$ in the $\text{PGCN}_g$ model.  
Similar to the $\text{PGCN}_n + \text{DropNode}$ model, we do not use dropout for $\text{PGCN}_g + \text{DropNode}$. 
For the $\text{PGCN}_g + \text{DropNode}$, the keep ratio $p$ is set to $p=0.75$, which is much higher than that used for the node classification models. This is due to the fact that in the considered graph classification datasets, the graph sizes are much smaller than those in the node classification datasets (see Table~\ref{table:graph_classification_dataset}). 
We train both models using a small learning rate of $0.0001$. 
\subsection{Experimental Results}
\label{sec:results}
\subsubsection{Node Classification}
\begin{table*}[t]
\centering
\caption{Node classification results in terms of the accuracy evaluation metric (\%). We report the mean and standard deviation of the accuracy over $100$ runs. The bold font indicates the best performance. Our models include $\text{PGCN}_n$ and $\text{PGCN}_n + \text{DropNode}$. In addition, we apply DropNode to the common GCN model~\citep{kipf2016semi}, referred to as $\text{GCN}_n + \text{DropNode}$. 
The asterisk (*) {\color{black}for DGCNN} indicates that the result is obtained by using our own implementation, {\color{black} where only the message passing formulation of the original method is employed.} 
}
\label{table:node_classification_result}
\footnotesize
 \begin{tabular}{c | c | c | c } 
 \hline \hline
 Method & CORA & CITESEER & PUBMED \\
\hline
\hline
GCN + DropEdge~\citeyearpar{rong2019dropedge} & $82.80$ & $72.30$ & $79.60$ \\
\hline
GMNN~\citeyearpar{qu2019gmnn}& $83.7$ & $72.9$ & $81.8$\\
\hline
$\text{GCN}$~\citeyearpar{kipf2016semi} & $81.9 \pm 0.7$ & $70.5 \pm 0.8$ & $78.9 \pm 0.5$\\
\hline
$\text{DGCNN}^*$~\citeyearpar{zhang2018}& $81.4\pm0.5$ & $69.8\pm0.7$ & $78.1\pm0.4$\\
\hline
GAT~\citeyearpar{velivckovic2017graph} & $83.0 \pm 0.7$ & $72.5 \pm 0.7$ & $79.0 \pm 0.3$\\
\hline
Graph U-Net~\citeyearpar{gao2019graph} & $84.4 \pm 0.6$ & $73.2 \pm 0.5$ & $79.6 \pm 0.2$\\
\hline
DGI~\citeyearpar{velivckovic2018deep} & $82.3\pm0.6$ & $71.8\pm0.
7$ & $76.8\pm0.6$\\
\hline \hline
$\text{PGCN}_n$ & $81.7\pm0.5$ & $70.6\pm0.7$ & $78.4\pm0.4$\\
\hline
$\text{GCN}_n + \text{DropNode}$ & $84.6\pm1.0$ & $74.3\pm0.5$ & $82.7\pm0.2$\\
\hline
$\text{PGCN}_n + \text{DropNode}$ & $\mathbf{85.1}\pm0.7$ & $\mathbf{74.3}\pm0.6$ & $\mathbf{83.0}\pm0.3$\\
\hline \hline
\end{tabular}
\end{table*}

The node classification results of different models are reported in~\tabref{table:node_classification_result}. The reported results include the mean and standard deviation of classification accuracy over $100$ runs. 
The results show that the $\text{PGCN}_n$ and GCN~\citep{kipf2016semi} models achieve higher accuracy compared to the DGCNN model. This can be  attributed to better message passing schemes giving smaller weights to popular nodes presented in \secref{sec:proposed_gcl} as these three models have similar configuration including number of graph convolutional layers and number of hidden units in each layer. 
Furthermore, the $\text{PGCN}_n$ model achieves marginally better performance compared to the popular GCN model on the CITESEER dataset, reaching $70.6$\% compared to $70.5$\% obtained by the GCN model. Nevertheless, the $\text{PGCN}_n$ model performs slightly worse than the GCN model on the CORA and PUBMED datasets, amounting to a $0.2$\% and $0.5$\% drop in terms of accuracy. 
{\color{black}
This can be explained by the fact that the diversity-balance coefficient ($\gamma$) of the CITESEER dataset is the highest (\ie $\gamma = 0.33$) compared to the CORA and PUBMED datasets (\ie $\gamma = 0.15$ and $\gamma = 0.13$, respectively, see~\tabref{table:node_classification_dataset}). 
The coefficient $\gamma$ reflects the diversity and balance of classes in the neighborhoods of high-degree nodes, and a higher value of $\gamma$ means that messages from high-degree nodes are not useful. 
Recall that the $\text{PGCN}_n$ model assigns  lower weights on high-degree nodes. Therefore, the $\text{PGCN}_n$ favors the CITESEER dataset compared to the others. 
}
In addition, compared to the other baseline models, the $\text{PGCN}_n$ model achieves lower accuracy on CORA, whereas it produces comparable results on CITESEER and PUBMED. 
When DropNode is used, it consistently improves the performance of all considered models. Specifically, $\text{PGCN}_n + \text{DropNode}$ and $\text{GCN}_n + \text{DropNode}$ significantly improve the performance of $\text{PGCN}_n$ and $\text{GCN}_n$ by around $4$ percentage points of accuracy. In particular, $\text{GCN}_n + \text{DropNode}$ can reach $84.6$\%, $74.3$\% and $82.7$\% while $\text{PGCN}_n + \text{DropNode}$ achieves the best performance with $85.1$\%, $74.3$\% and $83.0$\% on the CORA, CITESEER and PUBMED datasets, respectively. 
It is worth recalling that in our setting, the number of training examples is much smaller compared to the number of testing examples. By using DropNode, deformed versions of the underlying graph are created during each training epoch. In other words, DropNode acts as an augmentation technique on the training data which leads to an increased performance. 

\subsubsection{Graph Classification}
\begin{table*}[t]
\centering
\caption{Graph classification result in terms of percent (\%). FCN stands for fully-connected neural network ($2$ FC layers). 
N/A stands for not available. Daggers mean the results are produced by running the code of the authors on corresponding datasets (the results are not available in the original paper).
}
\label{table:graph_classification_result}
\footnotesize
\resizebox{\textwidth}{!}{%
 \begin{tabular}{c | c | c | c | c | c} 
 \hline \hline
 Method & PROTEINS & DD & ENZYMES & MUTAG & NCI1\\
\hline
\hline
Diff-Pool (GraphSAGE)~\citeyearpar{ying2018hierarchical} & $70.48$ & $75.42$ & $54.25$ & N/A & N/A\\
\hline
Diff-Pool (Soft Assign)~\citeyearpar{ying2018hierarchical} & $76.25$ & $80.64$ & $62.53$ & $88.89^\dagger$ & $80.36^\dagger$\\
\hline
Graph U-Net~\citeyearpar{gao2019graph} & $77.68$ & $\mathbf{82.43}$ & $48.33^\dagger$ & $86.76^\dagger$ & $72.12^\dagger$\\
\hline
CapsGNN~\citeyearpar{xinyi2018capsule} & $76.28$ & $75.38$ & $54.67$ & $86.67$ & $78.35$ \\
\hline
DGCNN~\citeyearpar{zhang2018} & $75.54$ & $79.37$ & $46.33^\dagger$ & $85.83$ & $74.44$\\
\hline
$\text{SAGPool}_g$~\citeyearpar{lee2019self} & $70.04$ & $76.19$ & N/A & N/A & $74.18$\\
\hline
$\text{SAGPool}_h$~\citeyearpar{lee2019self} & $71.86$ & $76.45$ & N/A & N/A & $67.45$\\
\hline
FCN ($2$FC) & $74.68$ & $75.47$ & $66.17$ & $87.78$ & $69.69$\\
\hline
$\text{GCN} + 2\text{FC}$ & $74.86$ & $75.64$ & $66.45$ & $86.11$ & $75.90$\\
\hline \hline
$\text{PGCN}_g+2\text{FC}$& $75.13$ & $78.46$ & $66.17$& $85.55$ & $75.84$\\
\hline
$\text{GCN}_g+\text{DropNode}$ & $76.58$ & $79.32$ & $69.00$ & $87.27$ & $79.03$\\
\hline
$\text{PGCN}_g+\text{DropNode}$ & $\mathbf{77.21}$ & $80.69$ & $\mathbf{70.50}$ & $\mathbf{89.44}$ & $\mathbf{81.11}$\\
\hline \hline
\end{tabular}
}
\end{table*}

The results for graph classification are given in Table~\ref{table:graph_classification_result}. In addition to the models mentioned in Section~\ref{sec:basic_model} and Section~\ref{sec:dropnode}, we also provide the result produced by a simple fully-connected neural network, denoted by FCN, with two hidden layers; each has size of $512$ units. 
We observe that the simple FCN can achieve high classification accuracy compared to presented strong baselines on some datasets. For instance, FCN obtains $74.68$\% on PROTEINS, which is around $4$ percentage points higher than the performance of GraphSAGE and $\text{SAGPool}_g$, and approximately $3$ percentage points higher than $\text{SAGPool}_h$. 
The good performance of  structure-blind fully-connect neural network has been reported by~\citep{luzhnica2019graph}, which is also confirmed in our work. 
By adding a GCONV or GPCONV layer on top of the FCN model ($\text{GCN} + 2\text{FC}$, $\text{PGCN}_g + 2\text{FC}$) the accuracy on PROTEINS, DD and ENZYMES is marginally improved while the accuracy on NCI$1$ is improved by $6$ percentage points. This is because the GCONV / GPCONV layers are able to exploit the graph structure of the considered bioinformatics datasets. 
{\color{black}
Additionally, the $\text{PGCN}_g + 2\text{FC}$ model favors the DD dataset as the coefficient $\gamma$ is the highest compared to other datasets, which is similar to node classification results (see~\tabref{table:graph_classification_dataset}). 
}
By using DropNode, the performance of our models is further improved. Specifically, our model with a single GPCONV layer, two FC layers and DropNode (\ie $\text{PGCN}_g+\text{DropNode}$) outperforms all the baselines on PROTEINS, ENZYMES, MUTAG and NCI1, except for DD where our models perform slightly worse compared to the Graph U-Net model. 
Even we do not outperform the Graph U-Net on the DD dataset, is is clear that DropNode improves the $\text{PGCN}_g$ by more than $2$\% accuracy point. 
This again confirms the consistency of DropNode in improving GCNN models.

\subsection{The Effect of DropNode on Deep GCNNs}
In this section, we investigate the effect of DropNode on deeper graph convolutional models. Specifically, we run our best model, which is comprised of many GPCONV layers with and without DropNode for node classification on the CORA and CITESEER datasets. The number of GPCONV layers $L_{\text{GPCONV}}$ is set to $5$, $7$ and $9$; each GPCONV layer has a hidden dimension of $64$. 
The numbers of nodes that are kept in each case are shown in~\tabref{table:node_threshold}. 
The rest of the parameters have the same values as presented in \secref{sec:hyperparameter_setting}. 
The corresponding results are presented in~\tabref{table:deep_gcn}.

We observe that the performance of the GCN and $\text{PGCN}_n$ models decreases significantly when the number of hidden layers increases. Specifically, $9$GPCONV-layer GCN produces an accuracy score of only $13$\% on CORA and $22.2$\% on CITESEER while similar performance is produced by a $\text{PGCN}_n$  model with $9$GPCONV layers. This can be explained by the fact that (i) deep GCN / $\text{PGCN}_n$ models have many more parameters compared to the shallow ones, which are prone to over-fitting, and (ii) the deep models suffer from over-smoothing~\citep{li2018deeper,chen2019measuring}, which results in indistinguishable node representations for the different classes. 
By applying DropNode on both models, the classification accuracy is improved significantly, especially in the case that $7$ and $9$ layers are used. This is because the effects of over-fitting and over-smoothing are alleviated. 

\begin{table}[t]
\centering
\caption{Number of nodes kept for downsampling layers. $3\text{GPCONV}$ indicates that there are three GPCONV layers used. 
\#DL stands for downsampling layer dimensionality. ``$-$'' means not applicable.}
\label{table:node_threshold}
\footnotesize
 \begin{tabular}{c | c | c | c | c } 
 \hline \hline
 & $3\text{GPCONV}$ & $5\text{GPCONV}$ & $7\text{GPCONV}$ & $9\text{GPCONV}$\\
 \hline
 \hline
  \#DL $1$ & $200$ & $200$ & $200$ & $200$ \\
  \hline
  \#DL $2$ & $-$ & $150$ & $150$ & $150$  \\
  \hline
  \#DL $3$ & $-$ & $-$ & $100$ & $100$  \\
  \hline
  \#DL $4$ & $-$ & $-$ & $-$ & $50$ \\
\hline \hline
\end{tabular}
\end{table}

\begin{table}[t]
\centering
\caption{Accuracy (\%) of deep GCNNs with DropNode integrated.}
\label{table:deep_gcn}
\footnotesize
\setlength\tabcolsep{1pt}
 \begin{tabular}{c|cc|cc|cc|cc} 
 \hline \hline
  & \multicolumn{2}{c|}{$3$ layers} & \multicolumn{2}{c|} {$5$ layers} & \multicolumn{2}{c|}{$7$ layers} & \multicolumn{2}{c}{$9$ layers} \\
  & CORA & CITESEER & CORA & CITESEER & CORA & CITESEER & CORA & CITESEER \\
  \hline
  \hline
  GCN & $79.1$ & $69.0$ & $78.8$ & $61.8$ & $46.2$ & $23.0$ & $13.0$ & $22.2$\\
  \hline
  PGCN & $79.8$ & $69.0$ & $77.6$ & $64.4$ & $52.7$ & $35.4$ & $13.0$ & $25.10$\\
 \hline
 \makecell{$\text{GCN}_n$ + \\ DropNode} & $84.60$ & $74.30$ & $80.80$ & $72.10$ & $71.80$ & $70.40$ & $\mathbf{51.20}$ & $66.70$ \\
 \hline
  \makecell{$\text{PGCN}_n$ + \\ DropNode} & $\mathbf{85.10}$ & $74.30$ & $\mathbf{81.40}$ & $\mathbf{72.30}$ & $\mathbf{75.10}$ & $\mathbf{70.70}$ & $49.60$ & $\mathbf{66.90}$ \\
\hline \hline
\end{tabular}
\end{table}

\subsubsection{The Effect of Dropping Strategies}
{\color{black}
To see the impact of the dropping strategy used in DropNode, we re-run the node classification experiments on the CORA, CITESEER and PUBMED datasets with $\text{PGCN}_g$ model and use the RW-dropping for DropNode. The results are presented in~\tabref{table:node_classification_sampling}. It can be seen that the RW-dropping-based DropNode leads to a better node classification performance compared to the B-dropping-based DropNode for all considered datasets. 
It is because the RW-dropping tends to produce small-world sub-graphs while the B-dropping tends to produce ER sub-graphs. The small-world sub-graphs have a high clustering coefficient, which statistically retains local topology of the original graph. On the other hand, the ER sub-graphs do not take into account the local structure of the original graphs, leading to a lower performance. 

In~\tabref{table:graph_property}, we present several statistical properties of the original graphs and sub-graphs including the average shortest path length (L), clustering coefficient (C) and a small-world coefficient typically denoted by $\sigma$~\citep{humphries2008network}; the sub-graphs are created by the B-dropping and RW-dropping approaches. These statistics are calculated for $100$ individual sub-graphs and we report the averaged numbers. We can see that sub-graphs created by B-dropping have small value for C, meaning that the number of triangles is small as the nodes are picked randomly. 
Inversely, sub-graphs produced by RW-dropping tends to preserve clusters, leading to much higher values for C and $\sigma$. 
In order to compute $\sigma$ for a sub-graph, an ensemble of $100$ equivalent ER graphs is generated. 
The average shortest length ($L_r$) and clustering coefficient ($C_r$) are computed for this ensemble, which are then used to calculate $\sigma$. 
Typically, a value of $\sigma$ greater than $1$ indicates the small-worldness of a graph.
These facts justify that the B-dropping and RW-dropping approaches tend to produce ER and small-world graphs, respectively.
}
\begin{table*}[t]
\centering
\caption{
{\color{black}
Node classification performance using B-dropping and RW-dropping approaches. 
}
}
\label{table:node_classification_sampling}
\footnotesize
 \begin{tabular}{c | c | c | c } 
 \hline \hline
 Method & CORA & CITESEER & PUBMED \\
\hline
$\text{PGCN}_n + \text{DropNode}$ (B-dropping) & ${85.1}\pm0.7$ & ${74.3}\pm0.6$ & ${83.0}\pm0.3$\\
\hline
$\text{PGCN}_n + \text{DropNode}$ (RW-dropping) & $\mathbf{85.9}\pm0.4$ & $\mathbf{74.9}\pm0.5$ & $\mathbf{83.3}\pm0.4$\\
\hline \hline
\end{tabular}
\end{table*}

\begin{table*}[t]
\centering
\caption{
{\color{black}
Graph properties with respect to different sampling strategies. 
L denotes the average shortest length, C denotes the clustering coefficient and $\sigma$ indicates the small-world coefficient. 
For sub-graphs, 
all criteria (L, C, and $\sigma$) are the average numbers calculated on a set of $100$ sub-graphs generated by the considered dropping strategies. 
The small-world coefficient $\sigma$ is not available for PUBMED because of the extremely computational expensiveness. 
}
}
\label{table:graph_property}
\footnotesize
 \begin{tabular}{c | c | c | c } 
 \hline \hline
 Dataset & Avg. Length (L) & Clustering Coeff. (C) & $\sigma$ \\
\hline
CORA & $6.31$ & $0.2407$ & $158.70$\\
\hline
CORA B sub-graphs & $1.81$ & $0.0086$ & $0.66$\\
\hline
CORA RW sub-graphs & $7.28$ & $0.2651$ & $14.72$ \\
\hline
CITESEER & $9.33$ & $0.1415$ & $190.18$ \\
\hline
CITESEER B sub-graphs & $1.47$ & $0.0021$ & $0.24$ \\
\hline
CITESEER RW sub-graphs & $5.63$ & $0.2111$ & $16.15$ \\
\hline
PUBMED & $6.34$ & $0.0602$ & N/A \\
\hline
PUBMED B sub-graphs & $1.79$ & $0.0011$ & $0.13$ \\
\hline
PUBMED RW sub-graphs & $8.43$ & $0.1163$ & $5.55$ \\
\hline \hline
\end{tabular}
\end{table*}


\section{Conclusion}
\label{sec:conclusion}
In this work, we have proposed a new graph message passing mechanism, which leverages the transition probabilities of nodes in a graph, for graph convolutional neural networks (GCNNs). 
The proposed message passing mechanism is simple, however, it achieves good performance for the common tasks of node and graph classification {\color{black} compared to several existing message passing formulations. 
We further showed that the proposed message passing tends to work well in the case that the neighborhoods of high-degree nodes are diversified and balanced and we proposed a coefficient to quantify the diversity and balance. }
Additionally, we have introduced a novel technique termed DropNode for regularizing the GCNNs. 
The DropNode regularization technique can be integrated into existing GCNN models leading to noticeable improvements for {\color{black} popular tasks (i.e., node and graph classification) on benchmark datasets. 
DropNode can be realized with several node dropping strategies such as B-dropping or RW leading to different types of random graphs such as ER or small-world graphs. We empirically showed that the RW dropping  outperforms the B-dropping. }
Furthermore, it has been shown that DropNode works well under the condition that the number of labelled examples is limited, which is useful in many real-life applications when it is normally hard and expensive to collect a substantial amount of labelled data. 
Our future work will focus on generalizing the proposed method on large graphs,~\eg reducing the computational cost of re-computing intermediate  {\color{black} aggregation coefficient matrices in case multiple downsampling layers are employed. 
Another direction for future work is to investigate different graph sampling techniques, which can be used for DropNode. }
In addition, as our method is general, it could be applied to a wide range of applications involving graph-structured data such as social media or Internet-of-Things data. 


\bibliography{references}

\begin{thebibliography}{51}
\expandafter\ifx\csname natexlab\endcsname\relax\def\natexlab#1{#1}\fi
\providecommand{\url}[1]{\texttt{#1}}
\providecommand{\href}[2]{#2}
\providecommand{\path}[1]{#1}
\providecommand{\DOIprefix}{doi:}
\providecommand{\ArXivprefix}{arXiv:}
\providecommand{\URLprefix}{URL: }
\providecommand{\Pubmedprefix}{pmid:}
\providecommand{\doi}[1]{\href{http://dx.doi.org/#1}{\path{#1}}}
\providecommand{\Pubmed}[1]{\href{pmid:#1}{\path{#1}}}
\providecommand{\bibinfo}[2]{#2}
\ifx\xfnm\relax \def\xfnm[#1]{\unskip,\space#1}\fi
\bibitem[{Atwood \& Towsley(2016)}]{atwood2016diffusion}
\bibinfo{author}{Atwood, J.}, \& \bibinfo{author}{Towsley, D.}
  (\bibinfo{year}{2016}).
\newblock \bibinfo{title}{Diffusion-convolutional neural networks}.
\newblock In {\it \bibinfo{booktitle}{Advances in neural information processing
  systems}\/} (pp. \bibinfo{pages}{1993--2001}).
\bibitem[{Battaglia et~al.(2016)Battaglia, Pascanu, Lai, Rezende
  et~al.}]{battaglia2016interaction}
\bibinfo{author}{Battaglia, P.}, \bibinfo{author}{Pascanu, R.},
  \bibinfo{author}{Lai, M.}, \bibinfo{author}{Rezende, D.~J.} et~al.
  (\bibinfo{year}{2016}).
\newblock \bibinfo{title}{Interaction networks for learning about objects,
  relations and physics}.
\newblock In {\it \bibinfo{booktitle}{Advances in neural information processing
  systems}\/} (pp. \bibinfo{pages}{4502--4510}).
\bibitem[{Bianchi et~al.(2019)Bianchi, Grattarola, Livi \&
  Alippi}]{bianchi2019graph}
\bibinfo{author}{Bianchi, F.~M.}, \bibinfo{author}{Grattarola, D.},
  \bibinfo{author}{Livi, L.}, \& \bibinfo{author}{Alippi, C.}
  (\bibinfo{year}{2019}).
\newblock \bibinfo{title}{Graph neural networks with convolutional arma
  filters}.
\newblock {\it \bibinfo{journal}{arXiv preprint arXiv:1901.01343}\/}, .
\bibitem[{Bruna et~al.(2013)Bruna, Zaremba, Szlam \& LeCun}]{bruna2013spectral}
\bibinfo{author}{Bruna, J.}, \bibinfo{author}{Zaremba, W.},
  \bibinfo{author}{Szlam, A.}, \& \bibinfo{author}{LeCun, Y.}
  (\bibinfo{year}{2013}).
\newblock \bibinfo{title}{Spectral networks and locally connected networks on
  graphs}.
\newblock {\it \bibinfo{journal}{arXiv preprint arXiv:1312.6203}\/}, .
\bibitem[{Chen et~al.(2019)Chen, Lin, Li, Li, Zhou \& Sun}]{chen2019measuring}
\bibinfo{author}{Chen, D.}, \bibinfo{author}{Lin, Y.}, \bibinfo{author}{Li,
  W.}, \bibinfo{author}{Li, P.}, \bibinfo{author}{Zhou, J.}, \&
  \bibinfo{author}{Sun, X.} (\bibinfo{year}{2019}).
\newblock \bibinfo{title}{Measuring and relieving the over-smoothing problem
  for graph neural networks from the topological view}.
\newblock {\it \bibinfo{journal}{arXiv preprint arXiv:1909.03211}\/}, .
\bibitem[{Chen et~al.(2018)Chen, Ma \& Xiao}]{chen2018fastgcn}
\bibinfo{author}{Chen, J.}, \bibinfo{author}{Ma, T.}, \& \bibinfo{author}{Xiao,
  C.} (\bibinfo{year}{2018}).
\newblock \bibinfo{title}{Fastgcn: fast learning with graph convolutional
  networks via importance sampling}.
\newblock {\it \bibinfo{journal}{arXiv:1801.10247}\/}, .
\bibitem[{Defferrard et~al.(2016)Defferrard, Bresson \&
  Vandergheynst}]{defferrard2016convolutional}
\bibinfo{author}{Defferrard, M.}, \bibinfo{author}{Bresson, X.}, \&
  \bibinfo{author}{Vandergheynst, P.} (\bibinfo{year}{2016}).
\newblock \bibinfo{title}{Convolutional neural networks on graphs with fast
  localized spectral filtering}.
\newblock In {\it \bibinfo{booktitle}{Advances in neural information processing
  systems}\/} (pp. \bibinfo{pages}{3844--3852}).
\bibitem[{DeVries \& Taylor(2017)}]{devries2017improved}
\bibinfo{author}{DeVries, T.}, \& \bibinfo{author}{Taylor, G.~W.}
  (\bibinfo{year}{2017}).
\newblock \bibinfo{title}{Improved regularization of convolutional neural
  networks with cutout}.
\newblock {\it \bibinfo{journal}{arXiv preprint arXiv:1708.04552}\/}, .
\bibitem[{Do et~al.(2019)Do, Nguyen, Tsiligianni, Aguirre, La~Manna, Pasveer,
  Philips \& Deligiannis}]{do2019matrix}
\bibinfo{author}{Do, T.~H.}, \bibinfo{author}{Nguyen, D.~M.},
  \bibinfo{author}{Tsiligianni, E.}, \bibinfo{author}{Aguirre, A.~L.},
  \bibinfo{author}{La~Manna, V.~P.}, \bibinfo{author}{Pasveer, F.},
  \bibinfo{author}{Philips, W.}, \& \bibinfo{author}{Deligiannis, N.}
  (\bibinfo{year}{2019}).
\newblock \bibinfo{title}{Matrix completion with variational graph
  autoencoders: Application in hyperlocal air quality inference}.
\newblock In {\it \bibinfo{booktitle}{IEEE International Conference on
  Acoustics, Speech and Signal Processing (ICASSP)}\/} (pp.
  \bibinfo{pages}{7535--7539}).
\bibitem[{Do et~al.(2017)Do, Nguyen, Tsiligianni, Cornelis \&
  Deligiannis}]{do2017multiview}
\bibinfo{author}{Do, T.~H.}, \bibinfo{author}{Nguyen, D.~M.},
  \bibinfo{author}{Tsiligianni, E.}, \bibinfo{author}{Cornelis, B.}, \&
  \bibinfo{author}{Deligiannis, N.} (\bibinfo{year}{2017}).
\newblock \bibinfo{title}{Multiview deep learning for predicting twitter users'
  location}.
\newblock {\it \bibinfo{journal}{arXiv preprint arXiv:1712.08091}\/}, .
\bibitem[{Duvenaud et~al.(2015)Duvenaud, Maclaurin, Iparraguirre, Bombarell,
  Hirzel, Aspuru-Guzik \& Adams}]{duvenaud2015}
\bibinfo{author}{Duvenaud, D.~K.}, \bibinfo{author}{Maclaurin, D.},
  \bibinfo{author}{Iparraguirre, J.}, \bibinfo{author}{Bombarell, R.},
  \bibinfo{author}{Hirzel, T.}, \bibinfo{author}{Aspuru-Guzik, A.}, \&
  \bibinfo{author}{Adams, R.~P.} (\bibinfo{year}{2015}).
\newblock \bibinfo{title}{Convolutional networks on graphs for learning
  molecular fingerprints}.
\newblock In \bibinfo{editor}{C.~Cortes}, \bibinfo{editor}{N.~D. Lawrence},
  \bibinfo{editor}{D.~D. Lee}, \bibinfo{editor}{M.~Sugiyama}, \&
  \bibinfo{editor}{R.~Garnett} (Eds.), {\it \bibinfo{booktitle}{Advances in
  Neural Information Processing Systems 28}\/} (pp.
  \bibinfo{pages}{2224--2232}).
\newblock \bibinfo{publisher}{Curran Associates, Inc.}
\bibitem[{Erd{\H{o}}s \& R{\'e}nyi(1960)}]{erdHos1960evolution}
\bibinfo{author}{Erd{\H{o}}s, P.}, \& \bibinfo{author}{R{\'e}nyi, A.}
  (\bibinfo{year}{1960}).
\newblock \bibinfo{title}{On the evolution of random graphs}.
\newblock {\it \bibinfo{journal}{Publ. Math. Inst. Hung. Acad. Sci}\/},  {\it
  \bibinfo{volume}{5}\/}(1), \bibinfo{pages}{17--60}.
\bibitem[{Feng et~al.(2020)Feng, Zhang, Dong, Han, Luan, Xu, Yang, Kharlamov \&
  Tang}]{feng2020graph}
\bibinfo{author}{Feng, W.}, \bibinfo{author}{Zhang, J.}, \bibinfo{author}{Dong,
  Y.}, \bibinfo{author}{Han, Y.}, \bibinfo{author}{Luan, H.},
  \bibinfo{author}{Xu, Q.}, \bibinfo{author}{Yang, Q.},
  \bibinfo{author}{Kharlamov, E.}, \& \bibinfo{author}{Tang, J.}
  (\bibinfo{year}{2020}).
\newblock \bibinfo{title}{Graph random neural networks for semi-supervised
  learning on graphs}.
\newblock {\it \bibinfo{journal}{Advances in Neural Information Processing
  Systems}\/},  {\it \bibinfo{volume}{33}\/}.
\bibitem[{Gao \& Ji(2019)}]{gao2019graph}
\bibinfo{author}{Gao, H.}, \& \bibinfo{author}{Ji, S.} (\bibinfo{year}{2019}).
\newblock \bibinfo{title}{Graph u-nets}.
\newblock In {\it \bibinfo{booktitle}{International Conference on Machine
  Learning}\/} (pp. \bibinfo{pages}{2083--2092}).
\bibitem[{Gao et~al.(2018)Gao, Wang \& Ji}]{gao2018large}
\bibinfo{author}{Gao, H.}, \bibinfo{author}{Wang, Z.}, \& \bibinfo{author}{Ji,
  S.} (\bibinfo{year}{2018}).
\newblock \bibinfo{title}{Large-scale learnable graph convolutional networks}.
\newblock In {\it \bibinfo{booktitle}{Proceedings of the 24th ACM SIGKDD
  International Conference on Knowledge Discovery \& Data Mining}\/} (pp.
  \bibinfo{pages}{1416--1424}).
\bibitem[{Gilmer et~al.(2017)Gilmer, Schoenholz, Riley, Vinyals \&
  Dahl}]{gilmer2017neural}
\bibinfo{author}{Gilmer, J.}, \bibinfo{author}{Schoenholz, S.~S.},
  \bibinfo{author}{Riley, P.~F.}, \bibinfo{author}{Vinyals, O.}, \&
  \bibinfo{author}{Dahl, G.~E.} (\bibinfo{year}{2017}).
\newblock \bibinfo{title}{Neural message passing for quantum chemistry}.
\newblock In {\it \bibinfo{booktitle}{Proceedings of the 34th International
  Conference on Machine Learning-Volume 70}\/} (pp.
  \bibinfo{pages}{1263--1272}).
\newblock \bibinfo{organization}{JMLR. org}.
\bibitem[{Gomez et~al.(2017)Gomez, Chiem \& Delvenne}]{gomez2017dynamics}
\bibinfo{author}{Gomez, L.~G.}, \bibinfo{author}{Chiem, B.}, \&
  \bibinfo{author}{Delvenne, J.-C.} (\bibinfo{year}{2017}).
\newblock \bibinfo{title}{Dynamics based features for graph classification}.
\newblock {\it \bibinfo{journal}{arXiv preprint arXiv:1705.10817}\/}, .
\bibitem[{Hamilton et~al.(2017)Hamilton, Ying \&
  Leskovec}]{hamilton2017inductive}
\bibinfo{author}{Hamilton, W.}, \bibinfo{author}{Ying, Z.}, \&
  \bibinfo{author}{Leskovec, J.} (\bibinfo{year}{2017}).
\newblock \bibinfo{title}{Inductive representation learning on large graphs}.
\newblock In {\it \bibinfo{booktitle}{Advances in Neural Information Processing
  Systems}\/} (pp. \bibinfo{pages}{1024--1034}).
\bibitem[{Henaff et~al.(2015)Henaff, Bruna \& LeCun}]{henaff2015deep}
\bibinfo{author}{Henaff, M.}, \bibinfo{author}{Bruna, J.}, \&
  \bibinfo{author}{LeCun, Y.} (\bibinfo{year}{2015}).
\newblock \bibinfo{title}{Deep convolutional networks on graph-structured
  data}.
\newblock {\it \bibinfo{journal}{arXiv preprint arXiv:1506.05163}\/}, .
\bibitem[{Humphries \& Gurney(2008)}]{humphries2008network}
\bibinfo{author}{Humphries, M.~D.}, \& \bibinfo{author}{Gurney, K.}
  (\bibinfo{year}{2008}).
\newblock \bibinfo{title}{Network ‘small-world-ness’: a quantitative method
  for determining canonical network equivalence}.
\newblock {\it \bibinfo{journal}{PloS one}\/},  {\it \bibinfo{volume}{3}\/}(4),
  \bibinfo{pages}{e0002051}.
\bibitem[{Kearnes et~al.(2016)Kearnes, McCloskey, Berndl, Pande \&
  Riley}]{kearnes2016molecular}
\bibinfo{author}{Kearnes, S.}, \bibinfo{author}{McCloskey, K.},
  \bibinfo{author}{Berndl, M.}, \bibinfo{author}{Pande, V.}, \&
  \bibinfo{author}{Riley, P.} (\bibinfo{year}{2016}).
\newblock \bibinfo{title}{Molecular graph convolutions: moving beyond
  fingerprints}.
\newblock {\it \bibinfo{journal}{Journal of computer-aided molecular
  design}\/},  {\it \bibinfo{volume}{30}\/}(8), \bibinfo{pages}{595--608}.
\bibitem[{Kersting et~al.(2016)Kersting, Kriege, Morris, Mutzel \&
  Neumann}]{KKMMN2016}
\bibinfo{author}{Kersting, K.}, \bibinfo{author}{Kriege, N.~M.},
  \bibinfo{author}{Morris, C.}, \bibinfo{author}{Mutzel, P.}, \&
  \bibinfo{author}{Neumann, M.} (\bibinfo{year}{2016}).
\newblock \bibinfo{title}{Benchmark data sets for graph kernels}.
\newblock \bibinfo{note}{\url{http://graphkernels.cs.tu-dortmund.de}}.
\bibitem[{Kipf \& Welling(2016)}]{kipf2016semi}
\bibinfo{author}{Kipf, T.~N.}, \& \bibinfo{author}{Welling, M.}
  (\bibinfo{year}{2016}).
\newblock \bibinfo{title}{Semi-supervised classification with graph
  convolutional networks}.
\newblock {\it \bibinfo{journal}{arXiv preprint arXiv:1609.02907}\/}, .
\bibitem[{Lee et~al.(2019)Lee, Lee \& Kang}]{lee2019self}
\bibinfo{author}{Lee, J.}, \bibinfo{author}{Lee, I.}, \& \bibinfo{author}{Kang,
  J.} (\bibinfo{year}{2019}).
\newblock \bibinfo{title}{Self-attention graph pooling}.
\newblock {\it \bibinfo{journal}{arXiv preprint arXiv:1904.08082}\/}, .
\bibitem[{Leskovec \& Faloutsos(2006)}]{leskovec2006sampling}
\bibinfo{author}{Leskovec, J.}, \& \bibinfo{author}{Faloutsos, C.}
  (\bibinfo{year}{2006}).
\newblock \bibinfo{title}{Sampling from large graphs}.
\newblock In {\it \bibinfo{booktitle}{Proceedings of the 12th ACM SIGKDD
  international conference on Knowledge discovery and data mining}\/} (pp.
  \bibinfo{pages}{631--636}).
\newblock \bibinfo{organization}{ACM}.
\bibitem[{Li et~al.(2018)Li, Han \& Wu}]{li2018deeper}
\bibinfo{author}{Li, Q.}, \bibinfo{author}{Han, Z.}, \& \bibinfo{author}{Wu,
  X.-M.} (\bibinfo{year}{2018}).
\newblock \bibinfo{title}{Deeper insights into graph convolutional networks for
  semi-supervised learning}.
\newblock In {\it \bibinfo{booktitle}{Thirty-Second AAAI Conference on
  Artificial Intelligence}\/}.
\bibitem[{Li et~al.(2015)Li, Tarlow, Brockschmidt \& Zemel}]{li2015gated}
\bibinfo{author}{Li, Y.}, \bibinfo{author}{Tarlow, D.},
  \bibinfo{author}{Brockschmidt, M.}, \& \bibinfo{author}{Zemel, R.}
  (\bibinfo{year}{2015}).
\newblock \bibinfo{title}{Gated graph sequence neural networks}.
\newblock {\it \bibinfo{journal}{arXiv preprint arXiv:1511.05493}\/}, .
\bibitem[{Luzhnica et~al.(2019)Luzhnica, Day \& Li{\`o}}]{luzhnica2019graph}
\bibinfo{author}{Luzhnica, E.}, \bibinfo{author}{Day, B.}, \&
  \bibinfo{author}{Li{\`o}, P.} (\bibinfo{year}{2019}).
\newblock \bibinfo{title}{On graph classification networks, datasets and
  baselines}.
\newblock {\it \bibinfo{journal}{arXiv preprint arXiv:1905.04682}\/}, .
\bibitem[{Monti et~al.(2019)Monti, Frasca, Eynard, Mannion \&
  Bronstein}]{monti2019fake}
\bibinfo{author}{Monti, F.}, \bibinfo{author}{Frasca, F.},
  \bibinfo{author}{Eynard, D.}, \bibinfo{author}{Mannion, D.}, \&
  \bibinfo{author}{Bronstein, M.~M.} (\bibinfo{year}{2019}).
\newblock \bibinfo{title}{Fake news detection on social media using geometric
  deep learning}.
\newblock {\it \bibinfo{journal}{arXiv preprint arXiv:1902.06673}\/}, .
\bibitem[{Niepert et~al.(2016)Niepert, Ahmed \& Kutzkov}]{niepert2016learning}
\bibinfo{author}{Niepert, M.}, \bibinfo{author}{Ahmed, M.}, \&
  \bibinfo{author}{Kutzkov, K.} (\bibinfo{year}{2016}).
\newblock \bibinfo{title}{Learning convolutional neural networks for graphs}.
\newblock In {\it \bibinfo{booktitle}{International conference on machine
  learning}\/} (pp. \bibinfo{pages}{2014--2023}).
\bibitem[{NIST(2016)}]{shannon_diversity}
\bibinfo{author}{NIST} (\bibinfo{year}{2016}).
\newblock \bibinfo{title}{Shannon diversity index}.
\newblock
  \bibinfo{note}{\url{https://www.itl.nist.gov/div898/software/dataplot/refman2/auxillar/shannon.htm}}.
\bibitem[{Ortega et~al.(2018)Ortega, Frossard, Kova{\v{c}}evi{\'c}, Moura \&
  Vandergheynst}]{ortega2018graph}
\bibinfo{author}{Ortega, A.}, \bibinfo{author}{Frossard, P.},
  \bibinfo{author}{Kova{\v{c}}evi{\'c}, J.}, \bibinfo{author}{Moura, J.~M.}, \&
  \bibinfo{author}{Vandergheynst, P.} (\bibinfo{year}{2018}).
\newblock \bibinfo{title}{Graph signal processing: Overview, challenges, and
  applications}.
\newblock {\it \bibinfo{journal}{Proceedings of the IEEE}\/},  {\it
  \bibinfo{volume}{106}\/}(5), \bibinfo{pages}{808--828}.
\bibitem[{Qi et~al.(2017)Qi, Liao, Jia, Fidler \& Urtasun}]{qi20173d}
\bibinfo{author}{Qi, X.}, \bibinfo{author}{Liao, R.}, \bibinfo{author}{Jia,
  J.}, \bibinfo{author}{Fidler, S.}, \& \bibinfo{author}{Urtasun, R.}
  (\bibinfo{year}{2017}).
\newblock \bibinfo{title}{3d graph neural networks for rgbd semantic
  segmentation}.
\newblock In {\it \bibinfo{booktitle}{Proceedings of the IEEE International
  Conference on Computer Vision}\/} (pp. \bibinfo{pages}{5199--5208}).
\bibitem[{Qu et~al.(2019)Qu, Bengio \& Tang}]{qu2019gmnn}
\bibinfo{author}{Qu, M.}, \bibinfo{author}{Bengio, Y.}, \&
  \bibinfo{author}{Tang, J.} (\bibinfo{year}{2019}).
\newblock \bibinfo{title}{Gmnn: Graph markov neural networks}.
\newblock {\it \bibinfo{journal}{arXiv preprint arXiv:1905.06214}\/}, .
\bibitem[{Quek et~al.(2011)Quek, Wang, Zhang \& Feng}]{quek2011structural}
\bibinfo{author}{Quek, A.}, \bibinfo{author}{Wang, Z.}, \bibinfo{author}{Zhang,
  J.}, \& \bibinfo{author}{Feng, D.} (\bibinfo{year}{2011}).
\newblock \bibinfo{title}{Structural image classification with graph neural
  networks}.
\newblock In {\it \bibinfo{booktitle}{2011 International Conference on Digital
  Image Computing: Techniques and Applications}\/} (pp.
  \bibinfo{pages}{416--421}).
\newblock \bibinfo{organization}{IEEE}.
\bibitem[{Rahimi et~al.(2015)Rahimi, Cohn \& Baldwin}]{rahimi2015twitter}
\bibinfo{author}{Rahimi, A.}, \bibinfo{author}{Cohn, T.}, \&
  \bibinfo{author}{Baldwin, T.} (\bibinfo{year}{2015}).
\newblock \bibinfo{title}{Twitter user geolocation using a unified text and
  network prediction model}.
\newblock {\it \bibinfo{journal}{arXiv preprint arXiv:1506.08259}\/}, .
\bibitem[{Rong et~al.(2019)Rong, Huang, Xu \& Huang}]{rong2019dropedge}
\bibinfo{author}{Rong, Y.}, \bibinfo{author}{Huang, W.}, \bibinfo{author}{Xu,
  T.}, \& \bibinfo{author}{Huang, J.} (\bibinfo{year}{2019}).
\newblock \bibinfo{title}{Dropedge: Towards deep graph convolutional networks
  on node classification}.
\newblock In {\it \bibinfo{booktitle}{International Conference on Learning
  Representations}\/}.
\bibitem[{Ronneberger et~al.(2015)Ronneberger, Fischer \&
  Brox}]{ronneberger2015u}
\bibinfo{author}{Ronneberger, O.}, \bibinfo{author}{Fischer, P.}, \&
  \bibinfo{author}{Brox, T.} (\bibinfo{year}{2015}).
\newblock \bibinfo{title}{U-net: Convolutional networks for biomedical image
  segmentation}.
\newblock In {\it \bibinfo{booktitle}{International Conference on Medical image
  computing and computer-assisted intervention}\/} (pp.
  \bibinfo{pages}{234--241}).
\newblock \bibinfo{organization}{Springer}.
\bibitem[{Sch{\"u}tt et~al.(2017)Sch{\"u}tt, Arbabzadah, Chmiela, M{\"u}ller \&
  Tkatchenko}]{schutt2017quantum}
\bibinfo{author}{Sch{\"u}tt, K.~T.}, \bibinfo{author}{Arbabzadah, F.},
  \bibinfo{author}{Chmiela, S.}, \bibinfo{author}{M{\"u}ller, K.~R.}, \&
  \bibinfo{author}{Tkatchenko, A.} (\bibinfo{year}{2017}).
\newblock \bibinfo{title}{Quantum-chemical insights from deep tensor neural
  networks}.
\newblock {\it \bibinfo{journal}{Nature communications}\/},  {\it
  \bibinfo{volume}{8}\/}(1), \bibinfo{pages}{1--8}.
\bibitem[{Sen et~al.(2008)Sen, Namata, Bilgic, Getoor, Galligher \&
  Eliassi-Rad}]{sen2008collective}
\bibinfo{author}{Sen, P.}, \bibinfo{author}{Namata, G.},
  \bibinfo{author}{Bilgic, M.}, \bibinfo{author}{Getoor, L.},
  \bibinfo{author}{Galligher, B.}, \& \bibinfo{author}{Eliassi-Rad, T.}
  (\bibinfo{year}{2008}).
\newblock \bibinfo{title}{Collective classification in network data}.
\newblock {\it \bibinfo{journal}{AI magazine}\/},  {\it
  \bibinfo{volume}{29}\/}(3), \bibinfo{pages}{93--93}.
\bibitem[{Simonovsky \& Komodakis(2017)}]{simonovsky2017dynamic}
\bibinfo{author}{Simonovsky, M.}, \& \bibinfo{author}{Komodakis, N.}
  (\bibinfo{year}{2017}).
\newblock \bibinfo{title}{Dynamic edge-conditioned filters in convolutional
  neural networks on graphs}.
\newblock In {\it \bibinfo{booktitle}{Proceedings of the IEEE conference on
  computer vision and pattern recognition}\/} (pp.
  \bibinfo{pages}{3693--3702}).
\bibitem[{Srivastava et~al.(2014)Srivastava, Hinton, Krizhevsky, Sutskever \&
  Salakhutdinov}]{srivastava2014dropout}
\bibinfo{author}{Srivastava, N.}, \bibinfo{author}{Hinton, G.},
  \bibinfo{author}{Krizhevsky, A.}, \bibinfo{author}{Sutskever, I.}, \&
  \bibinfo{author}{Salakhutdinov, R.} (\bibinfo{year}{2014}).
\newblock \bibinfo{title}{Dropout: a simple way to prevent neural networks from
  overfitting}.
\newblock {\it \bibinfo{journal}{The journal of machine learning research}\/},
  {\it \bibinfo{volume}{15}\/}(1), \bibinfo{pages}{1929--1958}.
\bibitem[{Teney et~al.(2017)Teney, Liu \& van Den~Hengel}]{teney2017graph}
\bibinfo{author}{Teney, D.}, \bibinfo{author}{Liu, L.}, \& \bibinfo{author}{van
  Den~Hengel, A.} (\bibinfo{year}{2017}).
\newblock \bibinfo{title}{Graph-structured representations for visual question
  answering}.
\newblock In {\it \bibinfo{booktitle}{Proceedings of the IEEE Conference on
  Computer Vision and Pattern Recognition}\/} (pp. \bibinfo{pages}{1--9}).
\bibitem[{Veli{\v{c}}kovi{\'c} et~al.(2017)Veli{\v{c}}kovi{\'c}, Cucurull,
  Casanova, Romero, Lio \& Bengio}]{velivckovic2017graph}
\bibinfo{author}{Veli{\v{c}}kovi{\'c}, P.}, \bibinfo{author}{Cucurull, G.},
  \bibinfo{author}{Casanova, A.}, \bibinfo{author}{Romero, A.},
  \bibinfo{author}{Lio, P.}, \& \bibinfo{author}{Bengio, Y.}
  (\bibinfo{year}{2017}).
\newblock \bibinfo{title}{Graph attention networks}.
\newblock {\it \bibinfo{journal}{arXiv preprint arXiv:1710.10903}\/}, .
\bibitem[{Veli{\v{c}}kovi{\'c} et~al.(2018)Veli{\v{c}}kovi{\'c}, Fedus,
  Hamilton, Li{\`o}, Bengio \& Hjelm}]{velivckovic2018deep}
\bibinfo{author}{Veli{\v{c}}kovi{\'c}, P.}, \bibinfo{author}{Fedus, W.},
  \bibinfo{author}{Hamilton, W.~L.}, \bibinfo{author}{Li{\`o}, P.},
  \bibinfo{author}{Bengio, Y.}, \& \bibinfo{author}{Hjelm, R.~D.}
  (\bibinfo{year}{2018}).
\newblock \bibinfo{title}{Deep graph infomax}.
\newblock {\it \bibinfo{journal}{arXiv preprint arXiv:1809.10341}\/}, .
\bibitem[{Watts \& Strogatz(1998)}]{watts1998collective}
\bibinfo{author}{Watts, D.~J.}, \& \bibinfo{author}{Strogatz, S.~H.}
  (\bibinfo{year}{1998}).
\newblock \bibinfo{title}{Collective dynamics of ‘small-world’networks}.
\newblock {\it \bibinfo{journal}{nature}\/},  {\it
  \bibinfo{volume}{393}\/}(6684), \bibinfo{pages}{440--442}.
\bibitem[{Xinyi \& Chen(2019)}]{xinyi2018capsule}
\bibinfo{author}{Xinyi, Z.}, \& \bibinfo{author}{Chen, L.}
  (\bibinfo{year}{2019}).
\newblock \bibinfo{title}{Capsule graph neural network}.
\newblock In {\it \bibinfo{booktitle}{International Conference on Learning
  Representations}\/}.
\bibitem[{Yao et~al.(2019)Yao, Mao \& Luo}]{yao2019graph}
\bibinfo{author}{Yao, L.}, \bibinfo{author}{Mao, C.}, \& \bibinfo{author}{Luo,
  Y.} (\bibinfo{year}{2019}).
\newblock \bibinfo{title}{Graph convolutional networks for text
  classification}.
\newblock In {\it \bibinfo{booktitle}{Proceedings of the AAAI Conference on
  Artificial Intelligence}\/} (pp. \bibinfo{pages}{7370--7377}).
\newblock volume~\bibinfo{volume}{33}.
\bibitem[{Ying et~al.(2018)Ying, You, Morris, Ren, Hamilton \&
  Leskovec}]{ying2018hierarchical}
\bibinfo{author}{Ying, Z.}, \bibinfo{author}{You, J.}, \bibinfo{author}{Morris,
  C.}, \bibinfo{author}{Ren, X.}, \bibinfo{author}{Hamilton, W.}, \&
  \bibinfo{author}{Leskovec, J.} (\bibinfo{year}{2018}).
\newblock \bibinfo{title}{Hierarchical graph representation learning with
  differentiable pooling}.
\newblock In {\it \bibinfo{booktitle}{Advances in Neural Information Processing
  Systems}\/} (pp. \bibinfo{pages}{4800--4810}).
\bibitem[{Zhang et~al.(2018)Zhang, Cui, Neumann \& Chen}]{zhang2018}
\bibinfo{author}{Zhang, M.}, \bibinfo{author}{Cui, Z.},
  \bibinfo{author}{Neumann, M.}, \& \bibinfo{author}{Chen, Y.}
  (\bibinfo{year}{2018}).
\newblock \bibinfo{title}{An end-to-end deep learning architecture for graph
  classification}.
\newblock In {\it \bibinfo{booktitle}{Thirty-Second AAAI Conference on
  Artificial Intelligence}\/}.
\bibitem[{Zhou et~al.(2018)Zhou, Cui, Zhang, Yang, Liu \& Sun}]{zhou2018graph}
\bibinfo{author}{Zhou, J.}, \bibinfo{author}{Cui, G.}, \bibinfo{author}{Zhang,
  Z.}, \bibinfo{author}{Yang, C.}, \bibinfo{author}{Liu, Z.}, \&
  \bibinfo{author}{Sun, M.} (\bibinfo{year}{2018}).
\newblock \bibinfo{title}{Graph neural networks: A review of methods and
  applications}.
\newblock {\it \bibinfo{journal}{arXiv preprint arXiv:1812.08434}\/}, .

\end{thebibliography}

\end{document}